\documentclass[10pt,twocolumn,letterpaper]{article}
\usepackage[disable]{todonotes} 
\usepackage[shortlabels,inline]{enumitem}
\usepackage[pagenumbers]{iccv} 

\input{includes/base}
\newcommand\MakeUppercaseGreek[1]{
  \begingroup
    \let\psi\Psi
    \let\omega\Omega
    \let\gamma\Gamma
    \MakeUppercase{#1}
  \endgroup}
\newcommand{\vectorsym}[1]{\bm{#1}}

\newcommand{\brackets}[1]{\left(#1\right)}

\newcommand{\round}[1]{\left\lfloor#1\right\rceil}

\newcommand{\matsym}[1]{\mathbf{#1}}
\newcommand{\squareb}[1]{\left[{#1}\right]}

\newcommand{\clip}[3]{\mathrm{clip}\brackets{#1,#2,#3}}

\newcommand{\vecop}[1]{\mathrm{Vec}\brackets{#1}}
\newcommand\blfootnote[1]{%
  \begingroup
  \renewcommand\thefootnote{}\footnote{#1}%
  \addtocounter{footnote}{-1}%
  \endgroup
}

\newcommand{\floor}[1]{\left\lfloor#1\right\rfloor}

\newcommand{\W}[0]{\matsym{W}}

\newcommand{\y}[0]{\vectorsym{y}}

\newcommand{\x}[0]{\vectorsym{x}}

\newcommand{\nds}[0]{N_{\mathcal{D}}}

\newcommand{\nout}[0]{d_{{\ell}}^{\y}}
\newcommand{\nin}[0]{d_{{\ell}}^{\x}}
\newcommand{\noutb}[0]{d_{\vectorsym{y}}}
\newcommand{\ninb}[0]{d_{\vectorsym{x}}}

\newcommand{\wlayerl}[0]{\matsym{W}_{\ell}}

\newcommand{\wlayerlv}[0]{\vectorsym{w}_{\ell}}
\newcommand{\arlayerl}[0]{\matsym{A}_{\ell}^{(r)}}
\newcommand{\brlayerl}[0]{\matsym{B}_{\ell}^{(r)}}

\newcommand{\artlayerl}[0]{\accentset{\textstyle\sim}{\matsym{A}}_{\ell}^{(r)}}
\newcommand{\brtlayerl}[0]{\accentset{\textstyle\sim}{\matsym{B}}_{\ell}^{(r)}}

\newcommand{\wrtlaeryl}[0]{\widetilde{\matsym{W}}_{\ell}}
\newcommand{\wrtlaeryi}[0]{\widetilde{\matsym{W}}}
\newcommand{\alayerl}[0]{\matsym{A}_{\ell}}
\newcommand{\blayerl}[0]{\matsym{B}_{\ell}}

\newcommand{\qp}[0]{{\phi}}
\newcommand{\qpa}[0]{\qp_{\alayerl}}
\newcommand{\qpb}[0]{\qp_{\blayerl}}
\newcommand{\qpw}[0]{\qp_{\wlayerl}}
\newcommand{\qpar}[0]{\qp_{\alayerl}^{(r)}}
\newcommand{\qpbr}[0]{\qp_{\blayerl}^{(r)}}

\newcommand{\net}[0]{\vectorsym{f}}
\newcommand{\ds}[0]{\mathcal{D}}

\newcommand{\indmp}[0]{\mathcal{I}_{\ell} }
\newcommand{\csetl}[0]{\mathcal{P}_{\ell} }
\newcommand{\bal}[0]{\bit_{\ell}^{\matsym{A}}}
\newcommand{\bbl}[0]{\bit_{\ell}^{\matsym{B}}}
\newcommand{\narl}[0]{\bal} 
\newcommand{\nbrl}[0]{\bbl} 

\newcommand{\rl}[0]{r_{\ell}}
\newcommand{\hesslayer}[0]{\matsym{D}_{\ell}}
\newcommand{\bit}[0]{b}
\newcommand{\wal}[0]{\bit_{\wlayerl}}

\newcommand{\nlayer}[0]{L}
\newcommand{\wsvd}[0]{\matsym{T}}

\newcommand{\localloss}[0]{\Lambda_{\ell}}

\newcommand{\hatwlayerl}[0]{\hat{\matsym{W}}_{\ell}}
\newcommand{\regfunc}[0]{\Omega}

\DeclareMathOperator*{\argmax}{argmax} 

\definecolor{iccvblue}{rgb}{0.21,0.49,0.74}
\usepackage[pagebackref,breaklinks,colorlinks,allcolors=iccvblue]{hyperref}

\newcommand{\name}[0]{MLoRQ}
\title{\name: Bridging Low-Rank and Quantization for Transformer Compression}

\author{
Ofir Gordon \quad Ariel Lapid \quad Elad Cohen \quad Yarden Yagil \quad Arnon Netzer \quad Hai Victor Habi\\
Sony Semiconductor Israel\\
{\tt\small \{ofir.gordon, ariel.lapid, elad.cohen, yarden.yagil, arnon.netzer, hai.habi\}@sony.com}
}

\newcounter{phase}[algorithm]
\newlength{\phaserulewidth}
\newcommand{\setphaserulewidth}{\setlength{\phaserulewidth}}

\setphaserulewidth{.7pt}

\newcommand{\phase}[1]{%
   \leavevmode\llap{\rule{\dimexpr\labelwidth+\labelsep}{\phaserulewidth}}\rule{\linewidth}{\phaserulewidth}
  \vspace{-3ex}
  \STATE\strut\refstepcounter{phase}\textit{Stage~\thephase~--~#1}
  \\
  \vspace{-1.25ex}\leavevmode\llap{\rule{\dimexpr\labelwidth+\labelsep}{\phaserulewidth}}\rule{\linewidth}{\phaserulewidth}}

\begin{document}

\maketitle
\begin{abstract}

Deploying transformer-based neural networks on resource-constrained edge devices presents a significant challenge.
This challenge is often addressed through various techniques, such as low-rank approximation and mixed-precision quantization.
In this work, we introduce Mixed Low-Rank and Quantization (MLoRQ), a novel method that integrates both techniques. 
MLoRQ employs a two-stage optimization process to determine optimal bit-width and rank assignments for each layer, adhering to predefined memory constraints.
This process includes: 
(i) an intra-layer optimization that identifies potentially optimal compression solutions out of all low-rank and quantization combinations;
(ii) an inter-layer optimization that assigns bit-width precision and rank to each layer while ensuring the memory constraint is met.
An optional final step applies a sequential optimization process using a modified adaptive rounding technique to mitigate compression-induced errors in joint low-rank approximation and quantization.
The method is compatible and can be seamlessly integrated with most existing quantization algorithms.
MLoRQ shows state-of-the-art results with up to 15\% performance improvement, evaluated on Vision Transformers for image classification, object detection, and instance segmentation tasks.
\end{abstract}
    
\section{Introduction}
 Transformer-based deep neural networks (DNN)\blfootnote{Preprint} \cite{vaswani2017attention,price2025probabilistic, bert, vit, deit, liu2021swin} have shown state-of-the-art performance in various domains and tasks, including computer vision~\cite{vit,deit, liu2021swin,carion2020end}, weather prediction~\cite{price2025probabilistic}, and natural language processing~\cite{bert}. However, transformer-based DNNs typically experience considerable challenges in terms of computational complexity, power consumption, and latency. This makes their deployment on edge devices with limited memory and computational power a challenging task. 
The literature presents various approaches to tackling this challenge, including quantization~\cite{gholami2021survey, adaround,eptq}, low-rank approximation
~\cite{fwsvd,tfwsvd,lpaf,rankdyna}, and pruning~\cite{dong2017learningprune, hap}, among others. 

Quantization emerged as a useful technique to reduce model complexity by converting the weight and activation tensors into low-bit-width representations.
Specifically, Post-Training Quantization (PTQ), which aims to do so with minor complexity and only a small calibration dataset, demonstrated advantages in compressing vision transformers using single-bit width quantization~\cite{optq,li2023repq,osplus,10839431erq}.
Other studies in convolution neural network (CNN)~\cite{uhlich2019mixed,habi2020hmq,hawqv2,10205417,pandey2023practical,koryakovskiy2023one} as well as transformers~\cite{xiao2023patch,liu2021post,lrpqvit}, demonstrate significant advantages by employing mixed precision quantization, where each layer can be assigned with a different bit-width, out of a set of candidates.

Another promising technique for compressing transformers is low-rank approximation, which can exploit the prevalence of fully connected operations in transformer-based DNNs. 
Early approaches \cite{denton2014exploiting} used singular value decomposition (SVD) to obtain low-rank alternatives for the original weights. Recent advances use weighted-SVD, guided by the activation covariance matrix~\cite{svdllm} or Fisher approximation~\cite{fwsvd} of the Hessian, to improve performance.

\begin{figure*}
    \centering
     \resizebox{\linewidth}{!}{
    \input{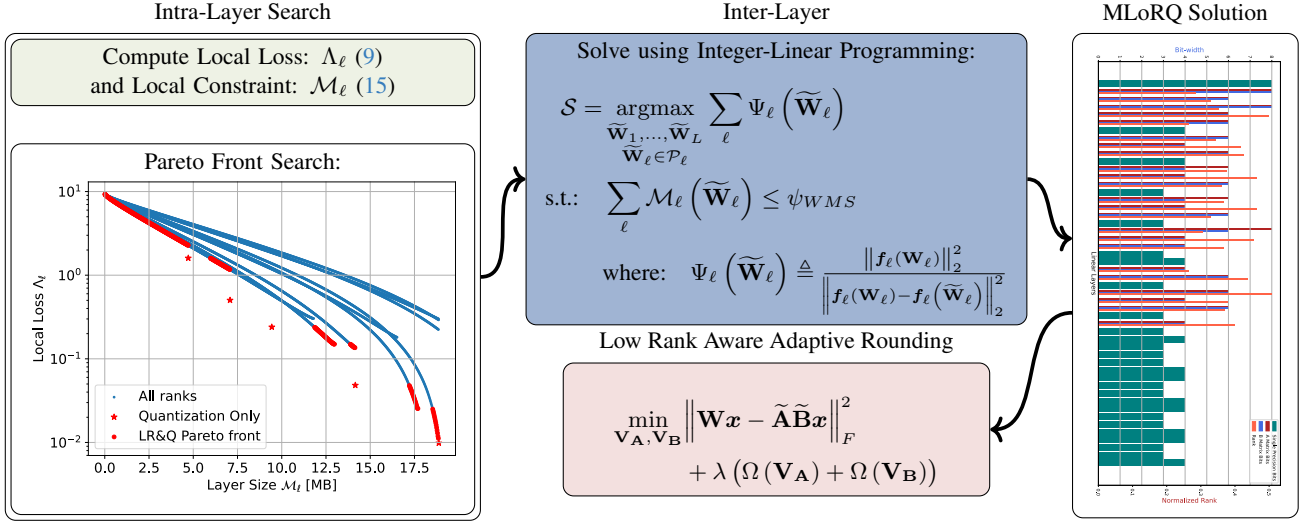}
    }
    \caption{The \textbf{\name{}} framework for Mixed Low-Rank and Quantization.
    \textbf{(i) Intra-layer} search (\ref{sec:intra-layer}) for the Pareto frontier optimal candidates set over the local loss and constraint. 
    \textbf{(ii) Inter-layer} search (\ref{sec:inter-layer}) to globally assign rank and bit-width to each layer. 
    \textbf{(iii)}  Apply complementary quantization algorithms with/without \textbf{low-rank-aware adaptive rounding}.
    }
    \label{fig:mlorp}
\end{figure*}

In this work, we propose a compression framework, targeted mainly for transformer-based models, named {\name}, which applies {joint}
Mixed Low-Rank and Quantization optimization.  
To the best of our knowledge,  this is the first attempt of \emph{joint} rank and quantization optimization.
We present a combined representation of these two techniques for a fully-connected layer.
Based on this representation, we introduce the mixed rank and precision problem, aiming to determine the smallest possible decrease in performance while adhering to memory constraints.
This presents a significant challenge, as the search space is considerably larger than when considering mixed precision or low-rank individually.

We address this challenge by proposing a two-stage algorithm.
First, an intra-layer search identifies a Pareto frontier of compression candidates, balancing compression and accuracy while significantly reducing the search space.
To further guide the optimization toward better solutions, we leverage Hessian knowledge, using the Label-Free Hessian approach~\cite{eptq}.
Then, an inter-layer optimization assigns bit-width and rank to each layer, selecting only from the filtered set of candidates obtained in the previous stage.
\name{} can be seamlessly integrated with most existing quantization schemes, enhancing them by incorporating both quantization and low-rank compression options.
Specifically, here we utilize the activation quantization enhancement from~\cite{li2023repq,10839431erq} with \name{}.
To further mitigate compression-induced error, we suggest an optional step of a modified adaptive rounding procedure~\cite{adaround} that considers both quantization and low-rank to refine the weight quantization rounding.

We conducted a series of numerical experiments to show the benefits of \name{} in computer vision---for the Image-Net classification task~\cite{russakovsky2015imagenet} and COCO~\cite{coco} object detection task---as well as in Natural Language Processing (NLP) on the General Language Understanding Evaluation (GLUE) benchmark~\cite{wang-etal-2018-glue}. 
\name{} outperforms existing PTQ techniques by strategically merging quantization and low-rank approximation, which is particularly beneficial at high compression rates.
For example, we show an improvement of up to 15\% in accuracy when compressing the ViT-B model to less than 12.5\% of its original size. 
In NLP, we achieve an improvement of up to 7\% when compressing BERT model~\cite{bert} weights.
Finally, we conducted a series of ablation studies to thoroughly analyze the contributions and benefits of each component in the proposed method.

Our contributions are summarized as follows:
\begin{enumerate}
    \item We introduce \name{}: a Mixed Low-Rank and Quantization approach enabling a notable reduction in the size and computational complexity of transformer-based DNNs.
    \item We propose a modified adaptive rounding technique for combined compression that includes both quantization and low-rank approximation.
    \item We present results for computer vision, including ImageNet classification and COCO object detection, where \name{} achieves state-of-the-art performance, demonstrating a notable improvement in accuracy over previous methods. 
\end{enumerate}

\section{Background \& Related Works}

\begin{figure*}[t]
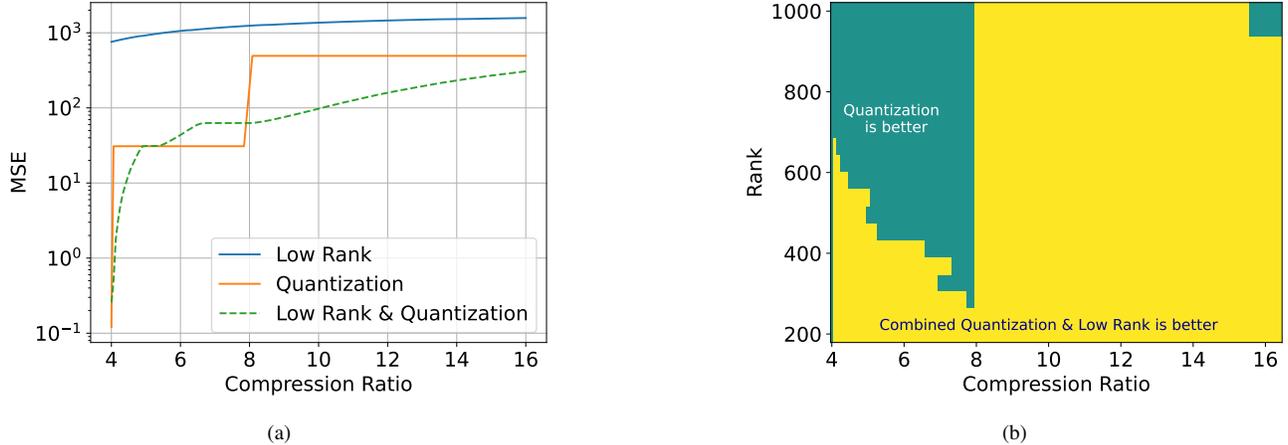

    \centering
    \begin{subfigure}{0.44\linewidth}
        \centering
        \includesvg[width=\textwidth]{images/motivation/rank_494.7368421052632.svg}
        \caption{}
        \label{fig:relative-comp}
    \end{subfigure}
    \hfill
    \begin{subfigure}{0.44\linewidth}
        \centering
 \includesvg[width=\textwidth]{images/motivation/all_ranks.svg}
    \caption{}
    \label{fig:best-repr}
    \end{subfigure}
    \caption{An analysis of different rank and compression rates: (a) Presenting the MSE of a single fully-connected layer under different relative compression ratios; and (b) Shows the best representation for a given rank and compression ratio.}
    \label{fig:analysis}
\end{figure*}

Here, we provide a short background on low-rank approximation and quantization methods utilized throughout this work. 
Due to space limitations, a detailed overview of related works is given in Appendix~\ref{sec:related-works}.

\subsection{Low Rank Approximation}

Low-rank approximation~\cite{srebro2003weighted} and tensor decomposition~\cite{golub1971singular,noach2020compressing} are powerful techniques for reducing neural network size and computational cost. 
In this study, we focus on compressing transformer-based neural networks, which primarily consist of fully-connected layers and $1\times1$ convolutions. 
This structure makes matrix low-rank approximation via singular value decomposition (SVD) a suitable choice for our approach.

Specifically, let $\noutb,\ninb\in\mathbb{N}$ and $\matsym{W}\in\mathbb{R}^{\noutb\times\ninb }$ represent a matrix for which we aim to reduce to rank $r \in \mathbb{N}/\{0\}$; 
its SVD and low-rank approximation form is given by
\begin{align}
\label{eq:svd_lr}
\matsym{W}&=\matsym{U}\matsym{\Sigma}\matsym{V}^T\approx\widetilde{\matsym{W}}=\matsym{U}^{\brackets{r}}\matsym{\Sigma}^{\brackets{r}}\brackets{\matsym{V}^{\brackets{r}}}^T
\end{align}
where $\matsym{U}$, $\matsym{V}$ are unitary matrices, and $\matsym{\Sigma}$ is a rectangular diagonal matrix containing the non-negative singular values of $\matsym{W}$.
The matrices $\matsym{U}^{\brackets{r}}$, $\matsym{V}^{\brackets{r}}$, and $\matsym{\Sigma}^{\brackets{r}}$ are reduced-rank versions of $\matsym{U}$, $\matsym{V}$ and $\matsym{\Sigma}$, respectively.

It has been demonstrated~\cite{fwsvd, svdllm} that this approach does not necessarily yield optimal neural network accuracy.
To address this, various methods have introduced additional weighted elements into the SVD. 
In the remainder of this section, we discuss weighted decomposition with a weighting factor $\wsvd$, where, for a given rank $r$, we aim to minimize the following objective:
\begin{align}
\label{eq:wsvd_problem}
\min\limits_{\matsym{A}^{(r)},\matsym{B}^{(r)}}\norm{\wsvd\brackets{\matsym{W}-\matsym{A}^{(r)}\matsym{B}^{(r)}}},
\end{align}
where $\matsym{A}^{(r)}\in\mathbb{R}^{\ninb \times r} \text{ and } \matsym{B}^{(r)}\in\mathbb{R}^{r \times \noutb}$ are the low-rank representation of the decomposed matrices of $\matsym{W}$.
\citet{fwsvd} and~\citet{srebro2003weighted} propose a solution to objective \eqref{eq:wsvd_problem} by employing SVD. 
Initially, SVD is performed on $\wsvd\matsym{W}$, yielding $\matsym{U}, \matsym{\Sigma}, \matsym{V} = svd\brackets{\wsvd\matsym{W}}$, where $svd\brackets{\matsym{M}}$ represents the singular value decomposition of matrix $\matsym{M}$. The resulting low-rank factorization is then provided by
$\matsym{A}^{(r)}=\wsvd^{-1}\matsym{U}^{(r)}\matsym{\Sigma}^{(r)}\quad \text{and}\quad \matsym{B}^{(r)}=\brackets{\matsym{V}^{(r)}}^T.$

\subsection{Quantization}
Quantization is central to neural network compression, with methods including log-quantization, power-of-two, uniform, symmetric, and more~\cite{gholami2021survey}.
In our work, we use a uniform quantizer, well supported by hardware. 
Let $\matsym{W}$ be a weight tensor to be quantized, with bit-width $\bit$, scaling parameter $s$, and zero point $z$.
The quantized weight is given by
\begin{equation*}
\mathrm{Q}\brackets{\matsym{W},\phi,b}\triangleq s\brackets{\clip{\round{\tfrac{\matsym{W}}{s}}+z}{0}{2^{b}-1}-z},
\end{equation*}
where $\phi=(s,z)$ denote the quantization parameters.
Various approaches have been suggested for choosing the scaling parameters $s$ and the zero point $z$, including mean square error (MSE)~\cite{nahshan2021loss}, min-max~\cite{krishnamoorthi2018quantizing}, and Hessian-MSE~\cite{eptq}. 
Recent studies introduced an additional optimization step to minimize rounding errors in weights quantization, specifically by choosing whether to round up or down. 
The majority of methods employ the soft-quantizer 
proposed by~\citet{adaround}:
\begin{align*}
    &Q_{S}\brackets{\matsym{W},\matsym{V},\phi,b}\triangleq\nonumber\\
    &s\brackets{\clip{\floor{\tfrac{\matsym{W}}{s}}+h\brackets{\matsym{V}}+z}{0}{2^{b}-1}- z},
\end{align*}
where $\matsym{V}\in\mathbb{R}^{\noutb\times\ninb}$ is an auxiliary variable used in the rounding optimization,  $h\brackets{x}\triangleq\clip{\sigma\brackets{x}\brackets{\xi-\gamma}+\gamma}{0}{1}$ is an element-wise function that ensures values remain between zero and one. Here, $\sigma$ represents the sigmoid function, and $\xi$ and $\gamma$ are stretch parameters with values $\xi=1.1$ and $\gamma=-0.1$. To facilitate the convergence of $h\brackets{\matsym{V}}$ to zero or one, a regularization term $\regfunc\brackets{\matsym{V}}\triangleq \sum_{i,j} 1-\abs{2 h\brackets{\matsym{V}}-1}^\beta$ is added into the objective loss using an annealing factor.



\begin{figure*}
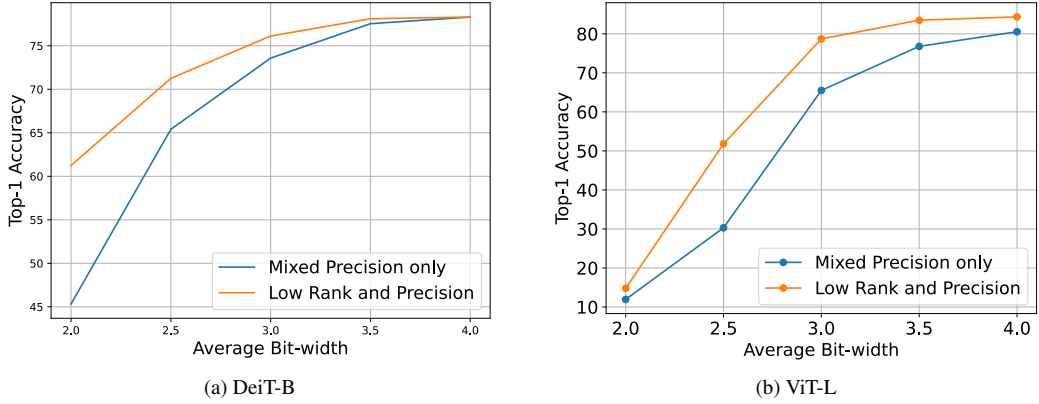

    \centering
    \begin{minipage}{0.8\linewidth}
        \centering
        \begin{subfigure}{0.48\linewidth}
            \centering
            \includesvg[width=\linewidth]{images/ablation_lr/ablation_lr_deit-b.svg}
            \caption{DeiT-B}
            \label{fig:lr-deitb}
        \end{subfigure}
        \hfill
        \begin{subfigure}{0.48\linewidth}
            \centering
            \includesvg[width=\linewidth]{images/ablation_lr/ablation_lr_vit-l.svg}
            \caption{ViT-L}
            \label{fig:lr-vitl}
        \end{subfigure}
    \end{minipage}
    \caption{Analysis of joint low-rank and quantization: 
    Comparison between joint optimization and mixed precision only on DeiT-B and ViT-L.}
    \label{fig:motivation}
\end{figure*}

\section{A Unified Low-Rank and Quantization Problem Definition}

We present a joint low-rank and quantization representation.
Specifically, let $\vectorsym{x}_{\ell}\in\mathbb{R}^{\nin}$, $\vectorsym{y}_{\ell}\in\mathbb{R}^{\nout}$ be the input and output vectors of the $\ell^{th}$ fully connected layer such that 
\begin{equation}
\label{eq:linear_operation}
    \vectorsym{y}_{\ell}=\wlayerl\vectorsym{x}_{\ell}+\vectorsym{b}_{\ell},
\end{equation}
where $\wlayerl\in\mathbb{R}^{\nout\times \nin}$, $\vectorsym{b}_{\ell}\in\mathbb{R}^{\nout}$ are the weight matrix and bias vector of the ${\ell}^{th}$ layer. Let $\wlayerl=\alayerl\blayerl$ be a decomposition of the layer's weights.
Given a rank~${\rl\in\mathcal{R}_{\ell}\triangleq[1,\min\brackets{\nin,\nout}]}$ and bit-widths~${\bal\in\mathcal{B}}$ and~${\bbl\in\mathcal{B}}$ for the decomposed matrices $\alayerl$ and $\blayerl$, respectively, where $\mathcal{B}$ is the set of bit-width options. 
We represent the compressed operation in \eqref{eq:linear_operation} as
\footnote{$\tilde{\vectorsym{x}}_{\ell}$ is the perturbed input of the $\ell^{th}$ layer as a result of the compression of preceding layers.}
\begin{equation}\label{eq:compressed_operation}
    \tilde{\vectorsym{y}}_{\ell}=\wrtlaeryl \tilde{\vectorsym{x}}_{\ell}+\vectorsym{b}_{\ell},
\end{equation}
where 
$\wrtlaeryl\in\mathcal{W}_{\ell}$ is the compressed weight.
Here, $\mathcal{W}_{\ell}$ is the set of compressed options of the layer $\ell^{th}$ that is given by:
\begin{align}\label{eq:compressed_set}
&\mathcal{W}_{\ell}=\mathcal{W}_{\ell}^{Q}\cup\mathcal{W}_{\ell}^{LQ},\quad\text{where}\\
    &\mathcal{W}_{\ell}^{Q}\triangleq\set{ Q\brackets{\wlayerl,\qp_{\wlayerl}\brackets{\wal},\wal}\mid \wal\in\mathcal{B}},\quad \text{and}\nonumber\\
    &\mathcal{W}_{\ell}^{LQ}\triangleq\set{ \artlayerl\brackets{\bal}\brtlayerl\brackets{\bbl} \mid \bal,\bbl\in\mathcal{B}, \rl\in\mathcal{R}_{\ell}}\nonumber   
\end{align}
are the sets of compressed weight by quantization only and joint quantization and low rank, respectively. 
In~\eqref{eq:compressed_set}, $\artlayerl\brackets{\bal}=Q\brackets{\matsym{A}_{\ell}^{(r)},\qpar\brackets{\bal},\bal}$ and $\brtlayerl\brackets{\bbl}=Q\brackets{\matsym{B}_{\ell}^{(r)},\qpbr\brackets{\bbl},\bbl}$ are quantized low-rank decomposition matrices, $\qpar, \qpbr$ and $\narl, \nbrl$ are the quantization parameters (scale and zero-point) and bit-width of the reduced rank matrices $\arlayerl$ and $\brlayerl$, respectively. 
It is important to note that \eqref{eq:compressed_set} is a \emph{hardware-friendly} operation, implementable by two consecutive fully-connected operations. 
In cases of a non-fully-connected operation (e.g., $3\times3$ convolution)~${\mathcal{W}_{\ell} = \mathcal{W}_{\ell}^{Q}}$. 

Considering the joint representation in \eqref{eq:compressed_operation}, our goal is to determine the rank and precision of each linear layer in the neural network while given a small representative dataset. This should result in minimal performance degradation while satisfying specified memory constraints. 
Specifically, let $\ds\in \set{\x^{(i)}_0}_{i=1}^{\nds}$ be a representative dataset that contains $\nds$ samples, $\net$ be a transformer-based neural network with $\nlayer$  fully-connected layers that has undergone pre-training with task loss $\mathcal{L}$. 
We formalize the problem as follows:
\begin{tcolorbox}
\begin{problem}
\label{problem}
{
Let $\net$  be a pre-trained neural network, 
$\mathcal{S}\triangleq\brackets{\wrtlaeryi_{1},\dots,\wrtlaeryi_{L}}$,
where $\wrtlaeryl$~${\in\mathcal{W}_{{\ell}}}$,
denotes the compressed weight assignment, $\Delta\vectorsym{w}$ is the vector representing the difference between the float and compressed weights.
Given the constraint of $\psi_{WMS}$ 
for the weights memory size, we give the following problem formulation:
\begin{align}
\label{eq:joint_problem}
\min_{\mathcal{S}}\quad&\mathcal{L}\brackets{\vectorsym{w}+\Delta\vectorsym{w}\brackets{\mathcal{S}}},\\
    \mathrm{s.t.:} \quad&\mathrm{WeightsMemory}\brackets{\mathcal{S}}\leq \psi_{WMS}, \nonumber
\end{align}
}
\end{problem}
\end{tcolorbox}

To motivate Problem~\ref{problem}, we analyze the MSE of a fully-connected layer for different ranks and compression ratios, from 25\% (8-bit) to 6.25\% (2-bit).
We evaluate three representations: quantization, low-rank, and the combined approach.
Figure~\ref{fig:analysis} (a) shows that none of the representations consistently outperforms the others.
Figure~\ref{fig:analysis} (b) shows which representation excels in different ranks and compression rates, highlighting that the combined approach performs best at low ranks and high compression. See details in Appendix~\ref{apx:motivation}.

\begin{figure*}
    \centering
    \resizebox{0.75\linewidth}{!}{
    \includesvg[width=0.8\linewidth]{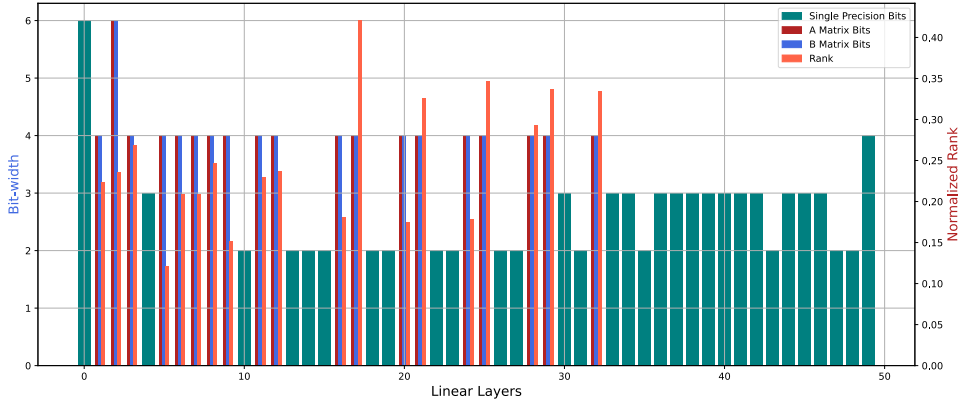}
    }
    \caption{The compression solution for the scenario of 2-bit average weights compression of DeiT-B with \name{}, where each layer is assigned with either a single precision bit-width or decomposed into low-rank matrices $A$ and~$B$.}
    \label{fig:comp-solution}
\end{figure*}

\section{ Mixed Low-Rank and Quantization}
\label{sec:mixed}
We derive the Mixed Low-Rank and Quantization ({\name}) method to solve the optimization problem stated in~\ref{problem}.
To address the large candidate space (as demonstrated in Table~\ref{tab:compare_methods_complexity} in Appendix~\ref{apx:search_space_table}), we split the search into two stages: 
\textbf{(i) Intra-layer} 
filtering using Pareto frontier optimization to find the set of potentially optimal candidates and drastically reduce the search space. 
\textbf{(ii) Inter-layer} 
search that assigns bit-widths and ranks for a unified compression solution.
After assigning a bit-width and rank to each layer, we include an optional stage that performs low-rank aware adaptive rounding, optimizing the weight quantization rounding values~\cite{adaround}.
In addition, we apply one of two activation quantization methods RepQ~\cite{li2023repq} or ERQ~\cite{10839431erq}, to correct quantization errors.  
We wish to highlight that \name{} can complement nearly any quantization method. The complete algorithm flow is depicted in Figure~\ref{fig:mlorp} and detailed in Algorithm~\ref{alg:workflow} in Appendix~\ref{apx:impl-detailes}.

To motivate the {\name} approach, we investigate the second-order Taylor approximation of the task loss~\cite{dong2019hawq,brecq}:
\begin{align}
    \label{eq:hessian-loss}
    \Delta\mathcal{L}= \vectorsym{g}^T \Delta\vectorsym{w} + \Delta\vectorsym{w}^T\matsym{H}_{\vectorsym{w}}\Delta\vectorsym{w}+\mathcal{O}\brackets{\Delta\vectorsym{w}^{3}},    
\end{align}
where $\Delta\mathcal{L}\triangleq\mathcal{L}\brackets{\vectorsym{w}}-\mathcal{L}\brackets{\vectorsym{w}+\Delta\vectorsym{w}}$ is the change in task loss due to compression, $\Delta\vectorsym{w}$ is the weights perturbation,  $\vectorsym{g}\triangleq \nabla_{\vectorsym{w}}\mathcal{L}\brackets{\vectorsym{w}}$ is the gradient of the task loss w.r.t. the weight vector $\vectorsym{w}$ and $\matsym{H}_{\vectorsym{w}}\triangleq\frac{\partial^2\mathcal{L}\brackets{\vectorsym{w}}}{\partial\vectorsym{w}\partial\vectorsym{w}^T}$ is the Hessian matrix of the task loss. 
Under the assumption that $\vectorsym{g}\approx 0$ as in \cite{adaround,brecq,eptq}, we get the following:
    $\Delta\mathcal{L}\approx  \Delta\vectorsym{w}^T\matsym{H}_{\vectorsym{w}}\Delta\vectorsym{w}$. 
    Using the assumption that the layers' contribution to the error induced by compression is independent as in \cite{adaround,hawqv2}, we have:
\begin{align}\label{eq:hessian_decomposition}
    \Delta\mathcal{L}\approx {\color{red}\overbrace{\sum_{\ell=1}^{L} \color{blue}{\underbrace{\Delta\vectorsym{w}^T_{\ell}\matsym{H}_{\vectorsym{w}_{\ell}}\Delta\vectorsym{w}  _{\ell}}_{\text{Intra Layer}}}}^{\text{Inter Layer}}}, 
\end{align}
where $\Delta\wlayerl=\wlayerl-\wrtlaeryl$ is the weight perturbation in matrix form,  $\Delta\vectorsym{w}_{\ell}=\vecop{\Delta\wlayerl}$ and $\matsym{H}_{\vectorsym{w}_{\ell}}$  are the weights perturbation and weights Hessian of the $\ell^{th}$ layer and $\vecop{\matsym{M}}$ is the vectorization of a matrix $\matsym{M}$. 
Equation~\ref{eq:hessian_decomposition}  illustrates the connection between weight perturbations caused by compression and task loss that we aim to reduce, thus forming the proposed solution to improve accuracy. According to \eqref{eq:hessian_decomposition},  the problem can be decomposed into distinct components. We begin with the intra-Layer stage, a phase dedicated to minimizing compression error within each layer.
Subsequently, in the latter stage, the objective shifts to minimizing the global error, which we refer to as the inter-layer stage. 
To obtain the Hessian matrix without labels, we utilize the Label-Free Hessian from~\cite{eptq}. 

\subsection{Intra-Layer Search}
\label{sec:intra-layer}
In this section, we perform the intra-layer search to determine a set of candidate solutions for the $\ell^{th}$ layer. 
This is achieved by minimizing the inner term in~\eqref{eq:hessian_decomposition}, under the added assumption that Hessian $\matsym{H}_{\vectorsym{w}_{\ell}}$ is a diagonal matrix. 
This results in the subsequent optimization problem:
\begin{equation}
\label{eq:problem_intra_final}
    \min\limits_{\arlayerl,\brlayerl,\qpa,\qpb}\localloss\brackets{\wrtlaeryl} \quad\forall \bal,\bbl, \rl
\end{equation}
where $\localloss\brackets{\wrtlaeryl}\triangleq \norm{\hesslayer\odot\brackets{\wlayerl-\wrtlaeryl}}_F^2$ is the local loss, $\hesslayer=\sqrt{\squareb{\matsym{H}_{\wlayerlv}}_{i+n\cdot j,i+n\cdot j}}$ represents the square root of the Hessian matrix's diagonal component for~$\wlayerl$,
and $\matsym{M}_1\odot\matsym{M}_2$ denotes the Hadamard product (element-wise multiplication) between matrices~$\matsym{M}_1$ and~$\matsym{M}_2$.
We solve problem~\eqref{eq:problem_intra_final} using a sequential method, starting with the low-rank decomposition, then selecting the quantization parameters, and ending with a search that provides us with all the potentially optimal solutions.

\subsubsection{Hessian-aware SVD}
We propose a Hessian-aware SVD applied to each layer's weight matrix $\matsym{W}_{\ell}$. 
Specifically, we solve the problem~\eqref{eq:problem_intra_final} without quantization for all ranks $r\in\mathcal{R}_{\ell}$, resulting in
\begin{align}\label{eq:problem_svd}
   \min\limits_{\arlayerl,\brlayerl} &\localloss\brackets{\arlayerl\brlayerl}  \quad \forall r\in\mathcal{R}_{\ell},
\end{align}
A closed-form solution to~\eqref{eq:problem_svd} is generally not available, as noted in \cite{srebro2003weighted}. 
Thus, to address problem~\eqref{eq:problem_svd}, we suggest minimizing an upper bound, which is given by
    $\localloss\brackets{\arlayerl\brlayerl}\leq
    \norm{\matsym{Q}_{\ell}\brackets{\wlayerl-\arlayerl\brlayerl}}_F^2$,
where $\squareb{\matsym{Q}_{\ell}}_{i,i}=\sum_{j}\squareb{\hesslayer}_{i,j}$. 
We start by applying an SVD on $\matsym{Q}_{\ell}\wlayerl$ to obtain $\matsym{U}_{\ell},\matsym{\Sigma}_{\ell},\matsym{V}_{\ell}=svd\brackets{\matsym{Q}_{\ell}\matsym{W}_{\ell}}$. 
Next, we combine the results of the Hessian-weighted SVD into two matrices. Given that weights are quantized using a scale per output channel, scaling individual output channels of the matrix does not affect quantization noise because the scaling is incorporated into the quantization scale parameter. 
Consequently, we select low-rank matrices so that the scaling can integrate into the output channel scale parameter without introducing additional noise, as given by
\begin{equation}
\label{eq:quant_aware_decomp}
\matsym{A}_{\ell}=\matsym{Q}_{\ell}^{-1}\matsym{U}_{\ell} \quad \text{and}\quad \matsym{B}_{\ell}=\matsym{\Sigma}_{\ell}\matsym{V}_{\ell}^T.
\end{equation}

\subsubsection{Quantization Parameters Selection}
Given $\wlayerl$, $\alayerl$, and $\blayerl$, we aim to find the quantization parameters $\qp$ (scale and zero-point) that minimize the quantization error for each matrix. 
For $\wlayerl$, the quantization parameters are determined for every bit-width using the Hessian-MSE method proposed in~\cite{eptq}, denoted by $\qp_{\wlayerl}$, which completes the construction of the set $\mathcal{W}^Q_{\ell}$.

Next, we address the quantization parameters of the decomposed matrices $\alayerl$ and $\blayerl$ to construct $\mathcal{W}^{LQ}_{\ell}$.
We aim to minimize while considering both the SVD and Hessian factorization to assess their impact on the task loss. 
To do this, we obtain the quantization parameters, $\qpa$ and $\qpb$, that minimize the following objective:
\begin{equation}\label{eq:miniziation_threshold}
\min\limits_{\qpa,\qpb}\localloss\brackets{\hatwlayerl}\quad\forall \bal,\bbl \in\mathcal{B}, 
\end{equation}
where
$\hatwlayerl=Q\brackets{\alayerl,\qpa,\bal}Q\brackets{\blayerl,\qpb,\bbl}$
is the quantized SVD without rank reduction. 
This means that we determine the quantization parameters once for all ranks. 
Given the large number of candidates, we propose solving~\eqref{eq:miniziation_threshold} separately for $\alayerl$ and $\blayerl$. 
For $\alayerl$, we solve the following problem:
\begin{equation}
\label{eq:threshold_upper_bound_a}   \qpa\brackets{\bit}=\arg\min\limits_{\qp}\localloss\brackets{Q\brackets{\alayerl,\qp,\bit}\blayerl}.
\end{equation}
As for $\blayerl$, we reduce the search complexity using the bound $\localloss\brackets{\alayerl Q\brackets{\blayerl,\qp,\bit}}\leq \norm{{\blayerl-Q\brackets{\blayerl,\qp,\bit}}}_F^2.$
This results in the following minimization problem:
\begin{equation}\label{eq:threshold_upper_bound_b}
     \qpb\brackets{\bit}=\arg\min\limits_{\qp}\norm{{\blayerl-Q\brackets{\blayerl,\qp,\bit}}}_F^2.
\end{equation}
The minimization of 
\eqref{eq:threshold_upper_bound_a} and
\eqref{eq:threshold_upper_bound_b} is performed on a fixed grid of points using a line search. 
\subsubsection{Pareto Set Search}
In the final step of this phase, we extract only relevant potentially optimal solutions for the $\ell^{th}$ layer 
{by identifying the set of Pareto optimal solutions.}
{
Specifically, let $\localloss\brackets{\wrtlaeryl}$ denote the loss~\eqref{eq:problem_intra_final}, 
and $\mathcal{M}_{\ell}\brackets{\wrtlaeryl}$ represents the memory footprint of the compressed weight, given by
\begin{equation}\label{eq:memory_constraint}
     \mathcal{M}_{\ell}\brackets{\wrtlaeryl}\triangleq 
     \begin{cases}
     \rl\brackets{ \nout \cdot \bal + \nin \cdot \bbl} &   \wrtlaeryl\in\mathcal{W}_{\ell}^{LQ}\\
     \nout\cdot\nin\cdot  \wal & \wrtlaeryl\in\mathcal{W}_{\ell}^{Q}.
     \end{cases}
\end{equation}
Let us define that a compression option
$\wrtlaeryl'$ \emph{dominates}~{($\succ$)} another option $\wrtlaeryl$ if both its local error and memory footprint are better and at least one of them is strictly better.
}
Then, a compression option is \emph{Pareto optimal} if no other compression option dominates it. 
Formally, the set of Pareto optimal solutions is denoted by~${\csetl=\set{\wrtlaeryl\in\mathcal{W}_{\ell} |\nexists\wrtlaeryl'\in\mathcal{W}_{\ell}:\wrtlaeryl'\succ\wrtlaeryl}}$. 
After obtaining $\localloss$ and $\mathcal{M}_{\ell}$ for all ranks and bit-widths, we select a set of candidate solutions that exist on the Pareto frontier as illustrated in the Pareto search section image in Figure~\ref{fig:mlorp}, and in Appendix~\ref{apx:pareto-ill}.

\subsection{Inter Layer Search}
\label{sec:inter-layer}
After acquiring the compression candidates $\csetl$ of each layer, We search for a mixed rank and quantization solution~$\mathcal{S}$ that assigns an optimal compression setup to each layer, minimizing the total error in the compressed model. 
While Hessian-based metrics have been widely used for global optimization~\cite{dong2019hawq,hawqv2,hap}, recent studies suggest instead to maximize the model's signal-to-quantization noise ratio (SQNR)~\cite{pandey2023practical,kim2023complexity} for this global optimization search. 
Our ablation study in Section~\ref{sec:exp-ablation} also demonstrates the superiority of this approach.
Thus, we formulate the problem as follows:
\begin{align}\label{eq:global_optimization}
    \mathcal{S}=&\argmax\limits_{\substack{{\wrtlaeryi_{1},\dots,\wrtlaeryi_{L}}\\
\wrtlaeryl\in\mathcal{P}_{\ell}\quad\forall\ell
}}
\sum_{\ell}\Psi_{\ell}\brackets{\wrtlaeryl}\\
\text{s.t.:}\quad&\sum_{\ell} \mathcal{M}_{\ell}\brackets{\wrtlaeryl} \leq \psi_{WMS}\nonumber
,
\end{align}
where 
$\Psi_{\ell}\brackets{\wrtlaeryl}\triangleq\tfrac{\norm{\net_{\ell}\brackets{\wlayerl}}_2^2}{\norm{\net_{\ell}\brackets{\wlayerl}-\net_{\ell}\brackets{\wrtlaeryl}}_2^2}$
is the SQNR of the $\ell^{th}$ layer perturbation, $\net_{\ell}$ is the output of a neural network as a function of the $\ell^{th}$ weight tensor, while the rest of the model is in floating point. 
For the cases where the set of $\bit_{\alayerl}$ and $\bit_{\blayerl}$ options have a large number of candidates for different ranks, we apply a linear interpolation {instead of running a full model inference,} to reduce the computational cost (see details in Appendix~\ref{apx:interpolation}). 
We translate the problem in \eqref{eq:global_optimization} into Integer Linear Programming (ILP) and use an off-the-shelf~\cite {ortools} algorithm as detailed in the Appendix~\ref{apx:mp}.

\begin{table*}[t]
\centering
\caption{
 Top-1 accuracy (\%) on ImageNet classification.
 \textbf{``+ RepQ''} and \textbf{``+ ERQ''} refer to combining \name{} with RepQ's activation parametrization and ERQ's activation error reduction techniques, respectively.
 \textbf{``+ LoRAda''} refers to adding our low-rank-aware adaptive rounding to each of the methods. \textbf{``DQ''} refers to ERQ's Dual Quantization technique~\cite{zhong2025towards}, incorporated into the method.
 All referenced results were taken from~\cite{10839431erq}.
}
\label{tab:full-vit-res}
\begin{tabular}{l|c|ccccccc}
\hline
\textbf{Model}                              & \textbf{\begin{tabular}[c]{@{}c@{}}Model\\ Size (\%) \end{tabular}}         & \textbf{ViT-S} & \textbf{ViT-B} & \textbf{DeiT-T} & \textbf{DeiT-S} & \textbf{DeiT-B} & \textbf{Swin-S} & \textbf{Swin-B} \\ \hline
\textbf{Full Prec.}                         & 100\%                & \textbf{81.39} & \textbf{84.54} & \textbf{72.21}  & \textbf{79.85}  & \textbf{81.8}   & \textbf{83.23}  & \textbf{85.27}  \\ \hline
AdaRound \cite{adaround}   & \multirow{5}{*}{\begin{tabular}[c]{@{}c@{}}9.3\%\\  (\raisebox{0.3ex}{\texttildelow}{3-bit}) \end{tabular}} & 11.04          & 4.72           & 36.05           & 33.56           & 62.50           & 68.12           & 53.92           \\
BRECQ \cite{brecq}         &                      & 4.97           & 1.25           & 29.23           & 18.58           & 40.49           & 66.93           & 53.38           \\
QDrop \cite{qdrop}         &                      & 9.77           & 11.87          & 17.85           & 30.27           & 61.12           & 73.47           & 74.33           \\
GPTQ \cite{optq}              &                      & 23.32          & 44.63          & 42.25           & 48.95           & 61.75           & 66.71           & 71.43           \\
PD-Quant \cite{pdquant}    &                      & 4.56           & 21.81          & 41.87           & 41.65           & 53.63           & 70.07           & 56.48           \\
\hdashline 

RepQ-ViT \cite{li2023repq} &        \multirow{3}{*}{\begin{tabular}[c]{@{}c@{}}9.3\%\\  (\raisebox{0.3ex}{\texttildelow}{3-bit}) \end{tabular}}                & 15.65          & 26.98          & 29.34           & 45.82           & 58.92           & 59.83           & 44.17           \\
\textbf{{\name} + RepQ (Ours)}                               &               & 17.65 & 48.16 & 44.38  & 52.32  & 68.41  & 77.90  & 77.01  \\ 
\textbf{{\name} + RepQ + LoRAda (Ours)}                         &               & 42.18 & \textbf{69.56} & \textbf{52.17}  & 63.23  & 76.03  & 78.43  & 78.70  \\ 
\hdashline 

ERQ \cite{10839431erq}     &           \multirow{5}{*}{\begin{tabular}[c]{@{}c@{}}9.3\%\\  (\raisebox{0.3ex}{\texttildelow}{3-bit}) \end{tabular}}            & 45.68          & 53.88          & 44.09           & 57.63           & 70.33           & 75.08           & 75.78           \\
ERQ DQ \cite{zhong2025towards}    &                      & 60.13           & 72.37          & \textbf{52.73}           & 69.09           & 76.47           & 78.12           & 79.98           \\
\textbf{{\name} + ERQ (Ours)}                                &                & 23.69 & 60.16 & 46.74 & 58.32  & 71.61  & 78.02  & 78.90  \\ 
\textbf{{\name} + ERQ + LoRAda (Ours)}                          &                & \textbf{46.55} & \textbf{68.20} & 51.38 & \textbf{64.44}  & \textbf{76.12}  & \textbf{79.27}  & \textbf{79.06}  \\ 
\textbf{{\name} + ERQ DQ + LoRAda (Ours)}                               &              &\textbf{60.69}          & \textbf{76.598}          & \textbf{54.014}           & \textbf{69.628}           & \textbf{77.804}           & \textbf{78.88}           & \textbf{80.386}               \\
\hline

AdaRound \cite{adaround}   & \multirow{7}{*}{\multirow{2}{*}{\begin{tabular}[c]{@{}c@{}}12.5\%\\  (\raisebox{0.3ex}{\texttildelow}{4-bit}) \end{tabular}} } & 63.09          & 70.51          & 55.65           & 69.24           & 75.20            & 76.05           & 78.12           \\
BRECQ \cite{brecq}        &                      & 11.31          & 3.03           & 38.41           & 32.89           & 59.10           & 68.40            & 56.51           \\
QDrop \cite{qdrop}         &                      & 17.77          & 21.72          & 31.65           & 35.79           & 65.47           & 78.92           & 80.49           \\
GPTQ \cite{optq}            &                      & 67.59          & 75.12          & 58.96           & 70.85           & 76.10           & 80.17           & 81.08           \\
PTQ4ViT \cite{ptq4vit}      &                      & 42.57          & 30.69          & 36.96           & 34.08           & 64.39           & 76.09           & 74.02           \\
APQ-ViT \cite{apqvit}       &                      & 47.95          & 41.41          & 47.94           & 43.55           & 67.48           & 77.15           & 76.48           \\
PD-Quant \cite{pdquant}    &                      & 32.64          & 34.86          & 58.50            & 64.85           & 60.06           & 77.04           & 75.84           \\
\hdashline 

RepQ-ViT \cite{li2023repq} &      \multirow{3}{*}{\multirow{2}{*}{\begin{tabular}[c]{@{}c@{}}12.5\%\\  (\raisebox{0.3ex}{\texttildelow}{4-bit}) \end{tabular}} }      & 65.05          & 68.48          & 57.43           & 69.03           & 75.61           & 79.45           & 78.32           \\
\textbf{{\name} + RepQ (Ours)}                               &                & 64.61 & 75.98 & 59.11  & 70.44  & 76.78  & 80.15           & 81.96  \\
\textbf{{\name} + RepQ + LoRAda (Ours)}                               &                & 69.11 & \textbf{79.13} & 61.48  & 72.49  & \textbf{78.61}  & 80.12           & 82.32  \\ 
\hdashline

ERQ \cite{10839431erq}    &            \multirow{3}{*}{\begin{tabular}[c]{@{}c@{}}12.5\%\\  (\raisebox{0.3ex}{\texttildelow}{4-bit}) \end{tabular}}           & 68.91          & 76.63          & 60.29           & 72.56           & 78.23           & 80.74  & 82.44           \\
\textbf{{\name} + ERQ (Ours)}                               &                & 63.80 & 78.43 & 60.26  & 72.18  & 77.99  & 80.61         & 82.61  \\
\textbf{{\name} + ERQ + LoRAda (Ours)}                               &                & \textbf{69.12} & 78.84 & \textbf{61.55}  & \textbf{72.80}  & 78.30  & \textbf{80.76}           & \textbf{82.63}  \\ \hline
\end{tabular}
\end{table*}

\section{Low-Rank Aware Adaptive Rounding}
In the concluding phase, we first apply activation quantization correction, such as reparameterization and log-two-sqrt quantization correction~\cite{li2023repq}. 
Furthermore, activation quantization correction is performed using Ridge regression from ERQ~\cite{10839431erq}. We wish to highlight that \name{} can complement nearly any quantization method.

Given the structure of compressed weights, as a result of the combination of low-rank and quantization, we propose a modified version of Adaptive Rounding~\cite{adaround}, referred to as \textbf{LoRAda}, tailored for scenarios involving low-rank and quantization considerations. 
The procedure encompasses a sequence of optimizations in which, for each layer, we aim to minimize both the error caused by weight perturbations and the propagation errors throughout the model. 
Suppose $\matsym{A}\in\mathbb{R}^{d_{\y}\times r}$ and $\matsym{B}\in\mathbb{R}^{r\times d_{\x}}$ represent the low-rank decomposition of a certain layer whose original weight is denoted by $\matsym{W}\in\mathbb{R}^{d_{\y}\times d_{\x}}$. 
We define two continuous variables, $\matsym{V}_{\matsym{A}}\in\mathbb{R}^{d_{\y}\times r}$ and $\matsym{V}_{\matsym{B}}\in\mathbb{R}^{r\times d_{\x}}$, which are optimized as part of the sequence-mean mechanism given by
\begin{align*}
    \min\limits_{\matsym{V}_{\matsym{A}},\matsym{V}_{\matsym{B}}}&\norm{\W \x - \widetilde{\matsym{A}}\widetilde{\matsym{B}}\x    }_F^2+\lambda\brackets{\regfunc\brackets{\matsym{V}_{\matsym{A}}}+\regfunc\brackets{\matsym{V}_{\matsym{B}}}},
\end{align*}
where $\widetilde{\matsym{A}}=Q_{S}\brackets{\matsym{A},\matsym{V}_{\matsym{A}},\qp_\matsym{A},\bit_\matsym{A}}$ and $\widetilde{\matsym{B}}=Q_{S}\brackets{\matsym{B},\matsym{V}_{\matsym{B}},\qp_\matsym{B},\bit_\matsym{B}}$ are the soft-quantized low-rank decomposition matrices.

 \section{Experiments}
\label{sec:exp}

\begin{table*}[t]
\centering
\caption{Extended comparison of Top-1 (\%) accuracy results for ImageNet classification against other PTQ methods that utilize mixed precision quantization for weights and activations. 
\textsc{mp} indicates mixed precision quantization.}
\label{tab:act-mp}
\resizebox{0.95\linewidth}{!}{
\begin{tabular}{@{}l|c|c|ccccccc@{}}
\toprule
\textbf{Model} & \textbf{\begin{tabular}[c]{@{}c@{}}Weights\\ Memory (\%) \end{tabular}} & \textbf{\begin{tabular}[c]{@{}c@{}}Activation \\ Bits \end{tabular}} & \textbf{ViT-S} & \textbf{ViT-B} & \textbf{DeiT-T} & \textbf{DeiT-S} & \textbf{DeiT-B} & \textbf{Swin-S} & \textbf{Swin-B} \\ \midrule
\textbf{Full Prec.} & \textbf{100\%} & \textbf{32} & \textbf{81.39} & \textbf{84.54} & \textbf{72.21} & \textbf{79.85} & \textbf{81.80} & \textbf{83.23} & \textbf{85.27} \\ \midrule
Liu \cite{liu2021post}                      & \multirow{3}{*}{\begin{tabular}[c]{@{}c@{}}12.5\%\\  (\raisebox{0.3ex}{\texttildelow}{4-bit}) \end{tabular}}   &    \multirow{2}{*}{4\textsc{mp}}       & -              & -              & -               & -               & 75.94           & -               & -                \\
LRP-QViT \cite{lrpqvit}                &     &               & 70.81          & 75.37          & 61.24           & 72.43           & 78.13           & 81.37           & 80.77                \\ 
\textbf{{\name} + ERQ + LoRAda (Ours)}                 &              &              &\textbf{75.28}          & \textbf{82.36}          & \textbf{66.17}           & \textbf{75.99}           & \textbf{78.81}           & \textbf{82.83}           & \textbf{84.80}               \\ \hline
PMQ \cite{xiao2023patch} & \multirow{2}{*}{\begin{tabular}[c]{@{}c@{}}12.5\%\\ 
(\raisebox{0.3ex}{\texttildelow}{4-bit}) \end{tabular}} & \multirow{2}{*}{8\textsc{mp}} & - & 68.91 & - & 73.69 & 78.85 & - & - \\
\textbf{{\name} + ERQ + LoRAda (Ours)} & &  & \textbf{80.02} & \textbf{84.46} & \textbf{70.73} & \textbf{79.11} & \textbf{80.57} & \textbf{82.82} & \textbf{84.93} \\ \midrule

Liu \cite{liu2021post} & \multirow{4}{*}{\begin{tabular}[c]{@{}c@{}}18.75\%\\ 
(\raisebox{0.3ex}{\texttildelow}{6-bit}) \end{tabular}} & \multirow{5}{*}{6\textsc{mp}} & - & 73.33 & - & 76.68 & 79.64 & - & - \\
PMQ & & & - & 75.26 & - & 75.10 & 77.47 & - & - \\
LRP-QViT \cite{lrpqvit} & & & 80.59 & 83.87 & 71.03 & 79.03 & 81.44 & 82.86 & 84.72 \\
AMP-ViT \cite{Tai_2025_WACV} & & & 79.90 & 83.12 & 71.10 & 79.29 & \textbf{81.29} & 63.08 & 60.18 \\
\textbf{{\name} + ERQ + LoRAda (Ours)} &  &  & \textbf{80.63} & \textbf{84.64} & \textbf{71.44} & \textbf{79.32} & 80.13 & \textbf{83.11} & \textbf{85.17} \\ \bottomrule
\end{tabular}%
}
\end{table*}

\begin{table*}[t]
\centering
\caption{Results on COCO dataset. 
``$\text{AP}^{\text{box}}$'' denotes the box average precision for object detection, and ``$\text{AP}^{\text{mask}}$'' denotes the mask average precision for instance segmentation.
All referenced results were taken from~\cite{zhong2025towards}.}
\label{tab:od}
\resizebox{0.95\linewidth}{!}{
\begin{tabular}{lccccccccc}
\hline
\multicolumn{1}{c}{\multirow{3}{*}{\textbf{Method}}} &
\multirow{3}{*}{\textbf{\begin{tabular}[c]{@{}c@{}}Model\\ Size (\%) \end{tabular}}} &
\multicolumn{4}{c}{Mask R-CNN} & \multicolumn{4}{c}{Cascade Mask R-CNN} \\ \cline{3-6}  \cline{7-10}
\multicolumn{2}{c}{} &
\multicolumn{2}{c}{\textbf{w. Swin-T}} & \multicolumn{2}{c}{\textbf{w. Swin-S}} &
\multicolumn{2}{c}{\textbf{w. Swin-T}} & \multicolumn{2}{c}{\textbf{w. Swin-S}} \\ 
\multicolumn{2}{c}{} &
$\text{AP}^{\text{box}}$ & $\text{AP}^{\text{mask}}$ & $\text{AP}^{\text{box}}$ & $\text{AP}^{\text{mask}}$ &
$\text{AP}^{\text{box}}$ & $\text{AP}^{\text{mask}}$ & $\text{AP}^{\text{box}}$ & $\text{AP}^{\text{mask}}$ \\
\hline
\textbf{Full-Precision} & \textbf{100\%} & \textbf{46.0} & \textbf{41.6} & \textbf{48.5} & \textbf{43.3} & \textbf{50.4} & \textbf{43.7} & \textbf{51.9} & \textbf{45.0} \\ \hline
GPTQ \cite{optq} & \multirow{5}{*}{\begin{tabular}[c]{@{}c@{}}9.3\%\\ (\raisebox{0.3ex}{\texttildelow}{3-bit}) \end{tabular}} & 22.9 & 25.0 & 31.7 & 32.5 & 39.8 & 35.8 & 44.6 & 39.6 \\
RepQ-ViT \cite{li2023repq} & & 22.2 & 24.1 & 27.9 & 29.9 & 40.2 & 35.7 & 43.7 & 38.8 \\
BRECQ \cite{brecq} & & 14.7 & 18.2 & 15.3 & 19.5 & 29.9 & 28.0 & 32.3 & 29.3 \\

ERQ \cite{zhong2025towards} & & 27.2 & 28.9 & 30.6 & 33.0 & 45.1 & 40.0 & 47.3 & 42.0 \\
\textbf{{\name} + ERQ + LoRAda (Ours)} & & \textbf{33.7} & \textbf{32.9} & \textbf{36.8} & \textbf{35.5} & \textbf{46.1} & \textbf{40.8} & \textbf{48.1} & \textbf{42.4} \\



\hline
\end{tabular}}
\end{table*}

We conducted experiments to demonstrate the advantages of \name{}. We first evaluated \name{} on the ImageNet dataset~\cite{russakovsky2015imagenet} using Vision Transformers (taken from timm~\cite{rw2019timm}), and on COCO with Mask R-CNN~\cite{he2017mask} and Cascade Mask R-CNN~\cite{cai2018cascade}, comparing it to leading compression methods. 
Subsequently, we tested \name{} on the General Language Understanding Evaluation (GLUE) benchmark~\cite{wang-etal-2018-glue} with BERT~\cite{bert}. 
Lastly, an ablation study was performed to gain further insight into the algorithmic components of \name{}.

Throughout our experiments, we applied the same setup unless otherwise specified. 
To ensure a fair comparison, we adopted quantization settings from prior work~\cite{li2023repq, 10839431erq}, using a uniform quantizer for weight and activation and a $log2$ quantizer for the softmax function.
We set the bit-width options set $\mathcal{B}$ to $\set{2,3,4,6,8}$.
Weights are quantized per output channel, and activations are per tensor.
A set of 1,024 images was used for the optimization process.
For the search of activation parameters and for the Hessian computation, we used a representative data set of 32 images. 
Additional implementation details and hyperparameters can be found in the Appendix~\ref{apx:impl-detailes}.

We report results in terms of the compressed weights memory size relative to the floating-point model weights memory (\%), where the compressed size accounts for both low-rank and quantization, as defined in~\eqref{eq:memory_constraint}.
This can be interpreted as an equivalent size to the weights quantized to some average bit-width precision.
Activations are quantized to uniform 4-bit in all experiments unless stated otherwise.

\subsection{Image Classification}
The results for the ViT-based classification task are presented in Table~\ref{tab:full-vit-res}. 
In this experiment, the model's weights are compressed using our \name{} method.
We incorporate three activation quantization enhancements into \name{}: RepQ's reparameterization~\cite{li2023repq} (\textbf{\name{} + RepQ}), ERQ's activation error reduction~\cite{10839431erq} (\textbf{\name{} + ERQ}), and ERQ's dual quantizer~\cite{zhong2025towards} (\textbf{\name{} + ERQ DQ}), highlighting \name{}'s standalone effectiveness, independent of dedicated rounding refinement.
Finally, we report results of our complete method, including our low-rank-aware adaptive rounding (\textbf{+ LoRAda}).

As discussed in this paper, {\name} is highly effective for compressing vision transformers due to their architecture, which leverages the advantages of low-rank compression.
The results presented in Table~\ref{tab:full-vit-res} support this claim and demonstrate {\name} advantages, especially for low-bit compression cases. 
{\name} achieves state-of-the-art results on all architectures, showing an improvement of up to 15\% on a wide range of models.

\paragraph{Activation Mixed Precision}
\begin{table*}[t]
\caption{GELU benchmark comparison on BERT-base language model for language understanding tasks. 
Compression is measured as the ratio of the compressed model's weights memory size to the full-precision model, following the methodology of the referenced works. The model's activations are kept in full-precision.}
\label{tab:bert}
\centering
\resizebox{\textwidth}{!}{
\begin{tabular}{l|c|c|ccccccc|c|c}
\hline
\multirow{2}{*}{\textbf{Method}} & \multirow{2}{*}{\textbf{\begin{tabular}[c]{@{}c@{}}Model\\ Size (\%) \end{tabular}}} & \textbf{CoNLL} & \textbf{MNLI} & \textbf{QNLI} & \textbf{MRPC} & \textbf{QQP} & \textbf{SST-2} & \textbf{CoLA} & \textbf{STS-B} & \multicolumn{1}{l|}{\textbf{G-Avg}} & \multicolumn{1}{l}{\textbf{A-Avg}} \\ \cline{3-12}
 & & \textbf{F1} & \textbf{Acc} & \textbf{Acc} & \textbf{F1} & \textbf{F1/Acc} & \textbf{Acc} & \textbf{Mcc} & \textbf{P/S Corr} & & \\ \hline
TFWSVD \cite{tfwsvd} & \multirow{4}{*}{16\%}  & 91.90 & 79.10 & 84.30 & 85.0 & 85.90 & 89.0 & 21.40 & 86.0 & 75.80 & 77.80 \\
 RankDyna \cite{rankdyna} & & 91.60 & 81.35 & 87.60 & 84.70 & 86.80 & 88.20 & 36.20 & 86.40 & 78.80 & 80.40 \\
 LPAF \cite{lpaf} & & - & 81.70 & 88.60 & 86.0 & - & 89.70 & 42.80 & - & - & - \\
 \textbf{{\name} (Ours)} & & \textbf{93.87} & \textbf{84.07} & \textbf{90.90} & \textbf{87.80} & \textbf{87.72/-} & \textbf{92.78} & \textbf{56.76} & \textbf{88.83} & \textbf{84.12} & \textbf{85.34} \\ \hline
 OS+* \cite{osplus} & \multirow{3}{*}{12.5\%} & - & 80.20 & 85.40 & 87.60 & 85.0/- & 91.40 & 50.0 & 86.50 & 80.90 & - \\
 OliVe \cite{olive} & & - & \textbf{84.10} & - & 86.76 & -/90.36 & 92.43 & 54.30 & - & - & - \\
 \textbf{{\name} (Ours)} & & \textbf{93.89} & 83.83 & \textbf{91.05} & \textbf{87.90} & \textbf{87.64/90.86} & \textbf{92.55} & \textbf{55.49} & \textbf{88.73} & \textbf{83.88} & \textbf{85.13} \\ \hline
TFWSVD \cite{tfwsvd} & \multirow{3}{*}{7.5\%} & 87.80 & 76.70 & 76.50 & 81.20 & 83.40 & 81.30 & 17.80 & 45.30 & 66.0 & 68.80 \\
 RankDyna \cite{rankdyna} & & 88.40 & 79.20 & 85.50 & 82.10 & 86.10 & 87.50 & 21.30 & 83.30 & 75.0 & 76.70 \\
 \textbf{{\name} (Ours)} & & \textbf{91.30} & \textbf{81.85} & \textbf{89.78} & \textbf{88.36} & \textbf{86.17/-} & \textbf{90.25} & \textbf{50.38} & \textbf{86.90} & \textbf{81.96} & \textbf{83.12} \\ \hline
\end{tabular}
}
\end{table*}
In addition, we compared \name{} with other works that suggested a mixed precision solution suited for transformer quantization.
For this, we incorporate the mixed activation precision in \name{}, which allows us to select different bit widths for activation quantization across layers, in addition to the discussed weights mixed rank and precision solution. 
The activation mixed-precision is solved separately following the approach detailed in~\cite{habi2020hmq} to constrain the maximal size of the feature map in the network~\cite{uhlich2019mixed}. Additional details are provided in Appendix~\ref{apx:act-mp}.
\name{} also excels in this setting, achieving better results compared to the other PTQ methods.

\subsection{COCO Object Detection}
We evaluated \name{} on the COCO dataset for the object-detection and instance-segmentation tasks on the Mask-RCNN~\cite{he2017mask} and Cascade Mask R-CNN~\cite{cai2018cascade} models.
The results are presented in Table~\ref{tab:od}.
\name{} shows improvement over state-of-the-art methods in different settings with over 6 percentage points in ``AP bbox'' for Mask R-CNN. 
This further highlights the effectiveness of combining quantization and low-rank approximation for compressing transformer-based models.

\subsection{GLUE Benchmark Evaluation}
We further evaluate {\name}'s ability to compress transformer-based models beyond vision tasks. 
To this end, we conducted experiments to assess the performance of a compressed BERT-base model~\cite{bert} on the GLUE benchmark~\cite{wang-etal-2018-glue}. 
For this, we performed per-task fine-tuning to prepare the BERT-base model for evaluating each task and then compressed the fine-tuned model.
The fine-tuning process and evaluation follow the details given by~\citet{bert}.
Table~\ref{tab:bert} compares {\name} with other PTQ methods designed to compress transformer-based language models. 
The table shows that the advantages of combining low-rank with mixed precision quantization extend effectively to language models, achieving state-of-the-art results with a significant margin on nearly all tasks and compression rates.

\subsection{Ablation Study}
\label{sec:exp-ablation}

\subsubsection{Joint Low-Rank and Quantization Ablation}
We evaluate the contribution of incorporating low-rank approximation with mixed precision. 
We compare the results of {\name} to the same algorithmic flow using only mixed precision options across various weight compression rates. 
Figure~\ref{fig:motivation} presents this comparison for the DeiT-B and the larger model ViT-L, with results for additional models in Appendix~\ref{apx:additional-results-lr}.
The results highlight the impact of integrating low-rank options into the optimization, particularly at high compression rates, where accuracy improvements are substantial.
This is shown in Figure~\ref{fig:comp-solution}, which illustrates the compression solution assigned by \name{} for the equivalence of a 2-bit weight compression, yielding significantly better results than its quantization-only counterpart.
Interestingly, for much smaller models like DeiT-T (Figure~\ref{fig:ablation-lr-deitt} in Appendix~\ref{apx:additional-results-lr}), incorporating low-rank options into the mixed-precision search has minimal effect.
This aligns with the understanding that low-rank approximation is more beneficial than quantization for large matrices and high compression rates.

\subsubsection{MLoRQ Components Ablation}
\label{apx:components-ablation}
Table~\ref{tab:ext-ablation} presents an additional ablation on each component of {\name}.
Each row in the table shows the results of {\name} with the default settings (Appendix~\ref{apx:impl-detailes}) when omitting one of the components or replacing it with a vanilla version, one that is not suited for a unified mixed-rank and quantization approach.
The bottom row shows the results for a complete execution of {\name}.

In the \emph{“Disabled Hessian-aware loss”} experiment, we remove the Hessian weights factor from the loss formulation in the intra-level optimization~\ref{sec:intra-layer}. 
The \emph{“Naive adaptive rounding”} experiment replaces the low-rank-aware adaptive rounding with a basic version that ignores the decomposed weight tensors. 
In the \emph{“Hessian-aware MP”} experiment, we substitute the inter-layer optimization global loss (Section~\ref{sec:inter-layer}) with an alternative loss for mixed precision optimization suggested by~\citet{hawqv2}. 
Each component independently enhances our method's performance, and their combination leads to the best results.




\begin{table}[t]
\centering
\caption{\name{} ablation study results.}
\label{tab:ext-ablation}
\resizebox{\linewidth}{!}{
\begin{tabular}{l|cccc}
\hline
\textbf{Mode}                                  & \textbf{ViT-S} & \textbf{DeiT-B} & \textbf{Swin-S} & \textbf{Swin-B} \\ \hline
Disabled Hessian-aware loss                                     & 39.64          & 76.1            & 78.11           & 77.04           \\
Naive adaptive rounding                        & 37.76          & 52.55           & 78.38           & 76.83           \\
Hessian-aware MP~\cite{hawqv2} & 29.73          & 70.04           & 75.62           & 76.09           \\
\textbf{\name}                          & \textbf{46.55} & \textbf{76.12}  & \textbf{78.74}  & \textbf{79.04}  \\ \hline
\end{tabular}%
}
\end{table}

\subsubsection{Number of Samples Ablation}

\begin{figure}[t]
    \centering    
    \includesvg[width=\linewidth]{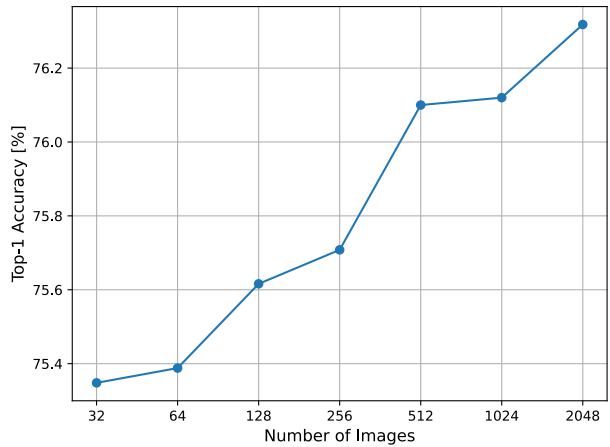}
    \caption{Number of samples ablation. Note that the number of samples axis is uniformly spaced for exposition purposes.}
    \label{fig:num-samples}
\end{figure}

We evaluate the effect of different numbers of calibration samples on the results of \name{} in Figure~\ref{fig:num-samples}.
As it can be seen, \name{} achieves a descent result even with a low number of images (32), and the gap between the chosen setting of 1,024 samples to an increased number is becoming marginal, making it a cost-effective choice.

\section{Conclusions}
In this study, we tackle joint low-rank approximation and mixed precision quantization by introducing \name{}, a novel algorithm that assigns rank and bit-width per layer. 
\name{} comprises two stages: 
(i) intra-layer candidates reduction via Pareto optimization; 
(ii) inter-layer rank and bit-width allocation.
The process is followed by an adaptive rounding refinement for mixed representations. 
It integrates easily into existing quantization schemes, unifying low-rank and quantization strategies. 
Experiments in vision tasks show notable improvement over state-of-the-art.
This work highlights the potential of combining quantization and low-rank approximation across tasks, with future directions including: 
(i) extending \name{} to large language models (LLMs) and vision-language models (VLMs) with high memory demands, and (ii) exploring additional compression techniques like pruning and one-bit quantization.

{\small
\bibliographystyle{ieeenat_fullname}
\bibliography{main}}
\clearpage
\appendix
\noindent\textbf{Table of Contents:}
\begin{itemize}
    \item Related Works can be found in Appendix~\ref{sec:related-works}.

    \item Extended details regarding the motivation for joint low-rank and quantization representation can be found in Appendix~\ref{apx:motivation}.

    \item Implementation details are provided in Appendix~\ref{apx:impl-detailes}.

    \item Search space analysis is provided in Appendix ~\ref{apx:search_space_table}.

    \item Optimization space trimming via Pareto Frontier can be found in Appendix~\ref{apx:interpolation}.

    \item Additional Low-Rank ablation results can be found in Appendix~\ref{apx:additional-results-lr}.

    \item Mixed solution illustration examples can be found in Appendix~\ref{apx:compression-solutions}.

    \item Additional Local Pareto Optimized Solutions in Appendix~\ref{apx:pareto-ill}.

    \item Extended motivation explanation for the joint problem approach is provided in Appendix~\ref{apx:seq-vs-joint}.
    
    \item Description of \name{} algorithmic flow  can be found in Appendix~\ref{apx:workflow}.

\end{itemize}
\section{Related Works}
\label{sec:related-works}

\subsection{Post-training Quantization for Transformers}

Neural network post-training quantization (PTQ) emerged as one of the most popular approaches for reducing model memory usage and computational complexity.
Many works from the last several years explored the approach of applying a short fine-tuning procedure to learn an adaptive rounding policy to improve PTQ performance, mainly for CNNs~\cite{adaround, brecq,qdrop,eptq}.
While performing well on convolutional neural network (CNN) models, these methods do not achieve comparable performance on transformer architectures due to their unique structure and different bottlenecks.
 Consequently, recent works have attempted to design dedicated PTQ methods for transformers~\cite{liu2021post, optq,ptq4vit,apqvit,noisyquant,osplus,olive}. 
 The main focus of these works was addressing the difficulties of quantizing the feature maps of Transformers, which tend to present massive outliers, making quantization a challenging task~\cite{bondarenko2024outliers,xiao2023smoothquant}.
 An important work in this direction, RepQ-ViT~\cite{li2023repq}, introduced a scale-reparameterized quantizer for post-LayerNorm and implemented a Log$\sqrt{2}$ quantizer to handle post-softmax activations quantization that mitigates outlier effects, leading to improved quantization robustness in Transformers.
 ERQ~\cite{10839431erq} extended this work and introduced a two-step PTQ approach to mitigate both the activation and the weight quantization error.
 Another approach, explored in several works, looks for a mixed-precision quantization solution that optimizes the trade-off between model compression and accuracy~\cite{hawqv2, xiao2023patch,  lrpqvit}.  These methods selectively assign lower bit widths to layers less sensitive to quantization.

\subsection{Low-rank Approximation for Model Compression}
Another method for reducing neural networks' memory and computational footprint is low-rank approximation, which decomposes the model's weight matrices into lower-rank representations.
Recently, this approach has attracted significant attention for transformer architectures due to their heavy reliance on large matrix multiplications~\cite{lpaf, aafm, yang2023efficient}.
FWSVD~\cite{fwsvd} and TFWSVD~\cite{tfwsvd} investigated various weighting strategies to enhance SVD decomposition for LR compression of Transformer-based language models.
RankDyna~\cite{rankdyna} introduced an iterative approach that dynamically assigns ranks based on the importance of the parameters, eliminating the need for a well-trained task-specific model.
Other works explored the benefits of LR primarily for large language model compression (LLM) and introduced different enhancements for these specific architectures~\cite{rankdyna,svdllm}.

\section{Local Error Analysis}
\label{apx:motivation}
\begin{figure*}
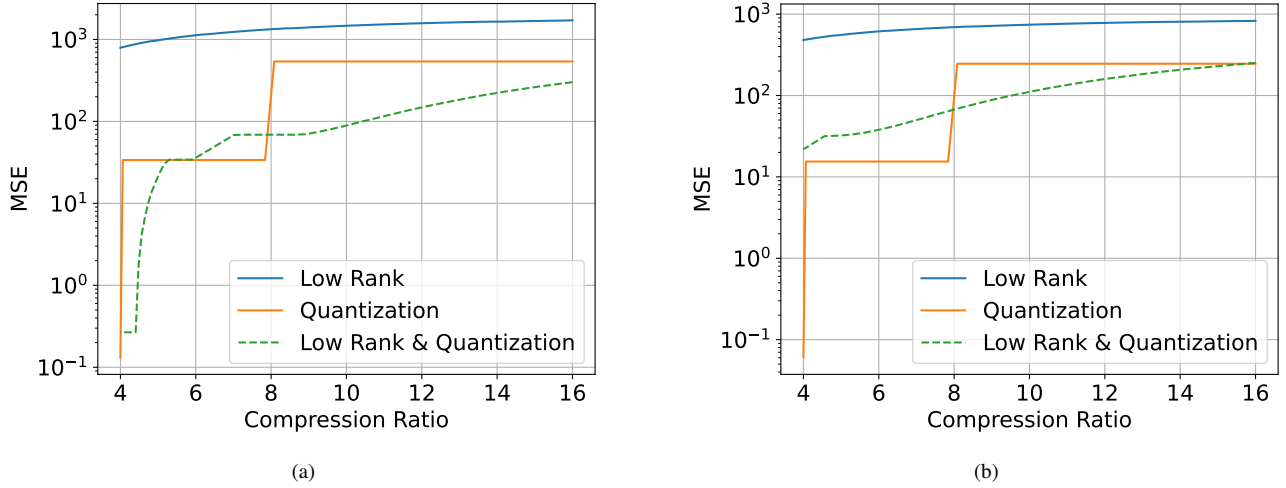

    \centering
    \begin{subfigure}{0.48\linewidth}
        \centering
        \includesvg[width=\textwidth]{images/motivation/rank_452.63157894736844.svg}
        \caption{}
        \label{fig:relative-comp-1}
    \end{subfigure}
    \hfill
    \begin{subfigure}{0.48\linewidth}
        \centering
     \includesvg[width=\textwidth]{images/motivation/rank_1000.0.svg}
    \caption{}
    \label{fig:relative-comp-2}
    \end{subfigure}
    \caption{Analysis of layer error over different compression rates.}
    \label{fig:mse_analysis}
\end{figure*}

To understand the advantages of integrating mixed rank and precision representations, we evaluate the loss in three compressed representations: low rank only, quantization only, and a combination of low rank and quantization. We study the compression of the matrix $\matsym{Q}=\matsym{M}_1\matsym{M}_2$ of rank $r$, with $\matsym{M}_1\in\mathbb{R}^{n\times r}$ and $\matsym{M}_2\in\mathbb{R}^{r\times m}$, both sampled from a standard Gaussian distribution. We then compressed $\matsym{Q}$ using three representations at compression rates ranging from 25\% (8bit) to {6.25\%} (2bit) and assessed the MSE. The quantization parameters were computed without clipping, and we used SVD for low-rank decomposition. The following figures show MSE for varying compression rates in ranks $r=453$ and $r=1000$.
We introduce a binary map illustrating optimal representations in varying compression rates and ranks, shown in Figure~\ref{fig:analysis}. 
Observe from Figure~\ref{fig:mse_analysis} that low-rank representations are ineffective below 25\% compression, though they provide advantages at larger ratios with minimal ranks.

\section{Implementation Details}
\label{apx:impl-detailes}


All experiments in Section~\ref{sec:exp} are performed in the following manner: first, we compute the Label-Free Hessian for all weights as shown in \cite{eptq} with 100 iterations and a batch size of 32 samples.  
In the following stage, we compute the decomposed matrices using Hessian-weighted SVD. 
For the quantization parameter selection, we conduct a detailed outlier search for each output channel of the weight using a percentile-based approach~\cite{10839431erq}. 
We evaluate the percentile values at $0.97, 0.98, 0.99, 0.995, 0.9995,$ $ 0.9997, 0.9999, 0.99995, 0.99999, 1$ and select the value that yields the best metrics in \eqref{eq:threshold_upper_bound_a} and \eqref{eq:threshold_upper_bound_b}. 
Moreover, in the case of quantization-only (without low-rank), we apply the same procedure with the Hessian MSE presented in~\cite{eptq}, which is given by
\begin{equation}\label{eq:hmse-loss}
   \qpw\brackets{\bit}=\arg\min\limits_{\qp}\norm{\hesslayer\odot\brackets{\wlayerl-Q\brackets{\wlayerl,\qp,\bit}}}_F^2.
\end{equation}

The parameter search is applied to every bit-width in the options set $\mathcal{B}$. 
Then, we compute the local loss \eqref{eq:problem_intra_final} and the local memory constraint \eqref{eq:memory_constraint}, which is used in the Pareto optimal solution search as in \cite{hawqv2}.
In the Inter-Layer Search stage, we wisely determine the bit-width and rank for every layer. 
{Next, to obtain the weight bit-width and rank, we apply a Integer-Linear Programming (ILP) optimization, which is detailed in Appendix~\ref{apx:mp}.}

 
During the final stage, we implement activation quantization correction using layer normalization parameterization correction and a logarithmic $\sqrt{2}$-quantizer, as outlined in \cite{li2023repq}. 
When integrating with ERQ, a Ridge Regression is introduced to further minimize the activation quantization error by adjusting the weights, according to \cite{10839431erq}.
Following the weighted adjustments for activation quantization, the Hessian, SVD, and quantization parameters are recomputed. 
Subsequently, sequential adaptive rounding optimization is employed to minimize the quantization rounding error by choosing whether to round up or down. 
We execute $20k$ gradient updates per layer using Adam with a learning rate of 0.3 (for ViT-S, the learning rate is 0.01) and a regularization factor of 0.3. 
All experiments were performed on an Nvidia A6000 GPU with a batch size of 32 images. 
The runtime of the algorithm to compress the ViT-S model is approximately one and a half hours.

\subsection{Mixed Precision Details}
\label{apx:mp}
\begin{figure*}[t]
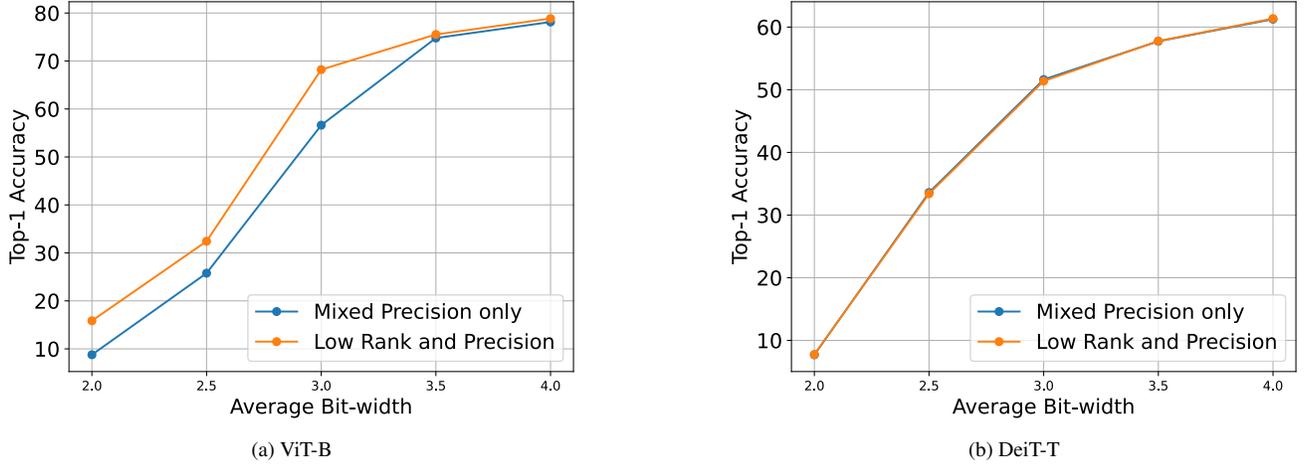

    \centering
    \begin{subfigure}{0.45\linewidth}
        \centering
        \includesvg[width=1.0\linewidth]{images/ablation_lr/ablation_lr_vit-b.svg}
    \caption{ViT-B}
    \label{fig:ablation-lr-vitb}
    \end{subfigure}
    \hfill
    \begin{subfigure}{0.45\linewidth}
        \centering
        \includesvg[width=1.0\linewidth]{images/ablation_lr/ablation_lr_deit-t.svg}
    \caption{DeiT-T}
    \label{fig:ablation-lr-deitt}
    \end{subfigure}
    \caption{A comparison of accuracy results when running compression with and without the low-rank option.}
\end{figure*}

To simplify the implementation process, rather than maximizing the SQNR, we minimize the normalized MSE.
as defined by
\begin{equation}
    \Phi_{\ell}\brackets{\wrtlaeryl}=\frac{1}{\Psi_{\ell}\brackets{\wrtlaeryl}}.
\end{equation}
Thus,  we formulate this problem as an Integer-Linear Programming (ILP) problem and use an off-the-shelf~\cite {ortools} algorithm to find a solution:
\begin{align}
\label{eq:ilp}
\min\limits_{\substack{{\wrtlaeryi_{1},\dots,\wrtlaeryi_{L}}\\
\wrtlaeryl\in\mathcal{P}_{\ell}\quad\forall\ell
}}\quad&\sum_{\ell}\sum_{\wrtlaeryl\in\csetl} \indmp\brackets{\wrtlaeryl} \cdot \Phi_{\ell}\brackets{\wrtlaeryl},\\
    \mathrm{s.t.} \quad&
    \sum_{\wrtlaeryl\in \csetl}\indmp\brackets{\wrtlaeryl} = 1,\nonumber \\
    \quad& \sum_{\ell}\sum_{\wrtlaeryl\in \csetl}\indmp\brackets{\wrtlaeryl}  \cdot \mathcal{M}_{\ell}\brackets{\wrtlaeryl} \leq \psi_{WMS},\nonumber \\
    \quad& \indmp\brackets{\wrtlaeryl} \in \set{0,1}, 
    \nonumber    
\end{align}

where $\indmp$ is an indicator of the select candidate of the $\ell^{th}$ layer.

\subsubsection{Activation Mixed Precision Details}
\label{apx:act-mp}
To comply with some of the evaluation comparisons presented in Section~\ref{sec:exp}, we incorporated a mixed precision activation search into our optimization framework.
For this, we utilize the maximal tensor size of activation memory as the activation memory constraint, which has a close-from solution as described in~\cite{habi2020hmq}, which is given by
\begin{equation}\label{eq:act_mp_close}
    b\brackets{\x_{\ell}}=\sup\set{b\mid b\leq\floor{\frac{\psi_{AWS}}{\mathrm{Size}\brackets{\x_{\ell}}}},b\in \mathcal{B}},
\end{equation}
where $Size\brackets{\x}$ denotes the size of the vector $\x$.

\section{Search Space Analysis}\label{apx:search_space_table}
In Table~\ref{tab:compare_methods_complexity}, we contrast the candidate solutions for a single fully connected layer as given in \eqref{eq:compressed_operation}. In the joint representation, complexity is quadratic with respect to bit-width options (typically 2-8) and linear with respect to the number of ranks (generally between 20-8000 in large language models). In most cases, the number of candidates is larger by an order of magnitude than low-rank and three times that of mixed precision.

\begin{table}
\centering
\caption{Compression of different compression approaches. $\mathcal{B}$ is the set of possible bit-widths, $\abs{\mathcal{B}}$ is the size of the set $\mathcal{B}$,  $b_A$, $r_{max}=min(\nout,\nin)$ is the maximal rank of the layer where $\nout$ and $\nin$ are the number of row and columns of the layer, respectively.}
\label{tab:compare_methods_complexity}
\resizebox{\linewidth}{!}{
\begin{tabular}{@{}lcccc@{}}
\toprule
\multicolumn{1}{c}{} & \begin{tabular}[c]{@{}c@{}}Floating\\ Pointg\end{tabular} & Low-Rank                       & Mixed Precision                     & \begin{tabular}[c]{@{}c@{}}Mixed Rank\\ \& Precision\end{tabular} \\ \midrule
Memory               &      $32\cdot\nout\cdot\nin      $                                                      &            $  32\cdot\brackets{\nout+\nin}\cdot \rl$                     &       $ b_{W}\cdot\nout\cdot\nin $                         &    $\brackets{\nout b_{A}+\nin b_{B}}\cdot \rl $                                                                         \\
Candidate            & 1                                     & $r_{max}$ & $|\mathcal{B}|$ & \multicolumn{1}{l}{$\abs{\mathcal{B}}\brackets{1+\abs{\mathcal{B}} r_{max}}$}                               \\ \bottomrule
\end{tabular}%
}
\end{table}
\section{Inter-Intra Layer Metric Interpolation}
\label{apx:interpolation}

Calculating the error directly using~\eqref{eq:global_optimization} for every configuration and layer poses a significant computational challenge, as each calculation requires independent inference on the entire calibration dataset, which can be costly in cases where the set $\mathcal{P}_{\ell}$ is very large. 

To reduce the computational cost in these cases, we suggest approximating the error by linear interpolation to overcome this limitation. This significantly reduces computational effort while maintaining accuracy.

\begin{figure}
    \centering    
    \includesvg[width=\linewidth]{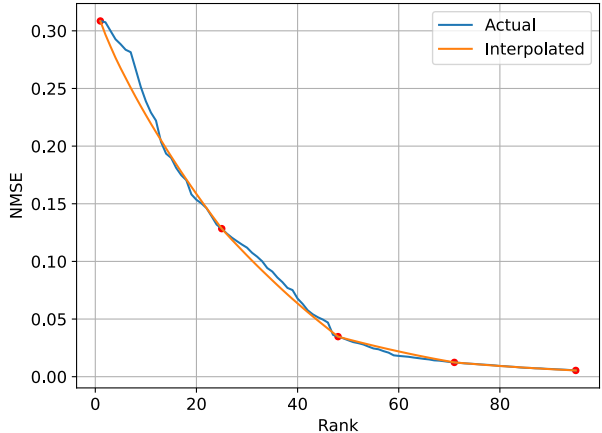}
    \caption{Illustration of the actual per-rank error value and the interpolation approximation error value on layer \emph{blocks.1.attn.proj} of model DeiT-T.}
    \label{fig:interpolation}
\end{figure}

In particular, we establish a computational budget of $k_{inf}$ metric evaluations for each bit-width pair $(\bal,\bbl)$. 
First, constrain the set of candidates $\mathcal{K}_{\ell}\brackets{b_A,b_B}=\set{\wrtlaeryl|n=b_A, m=b_B : \wrtlaeryl\in\mathcal{P}_{\ell}}$ 
with bit widths $b_A$, $b_B$ for the matrices $\alayerl$, $\blayerl$. 
If the size of $\mathcal{K}_{\ell}\leq k_{inf}$, compute the metric for each candidate. Alternatively, choose $k_{inf}$ so that these sizes are uniformly distributed between $\min\limits_{\wrtlaeryl\in\mathcal{K}_{\ell}\brackets{b_A,b_B}}\mathcal{M}_{\ell}\brackets{\wrtlaeryl}$ and $\max\limits_{\wrtlaeryl\in\mathcal{K}_{\ell}\brackets{b_A,b_B}}\mathcal{M}_{\ell}\brackets{\wrtlaeryl}$, resulting in the set $\mathcal{G}_{\ell}\brackets{b_A,b_B}$. Then the interpolated metric is given by
\begin{align}\label{eq:linear_interpolation}
    &\overline{\Phi}_{\ell}\brackets{\wrtlaeryl}\\
    &=\begin{cases}
        \Phi_{\ell}\brackets{\wrtlaeryl}  & \wrtlaeryl\in\mathcal{G}_{\ell}\brackets{b_A,b_B}\\
        \Phi_{\ell}\brackets{ \wrtlaeryl^{(l)}}\cdot\alpha+\Phi_{\ell}\brackets{\wrtlaeryl^{(h)}}\cdot \beta & \wrtlaeryl\notin\mathcal{G}_{\ell}\brackets{b_A,b_B}
    \end{cases}\nonumber
\end{align}
where $\wrtlaeryl^{(l)}, \wrtlaeryl^{(h)} \in \mathcal{G}_{\ell}\brackets{b_A,b_B}$ represent the lower and upper compressed weight around the current compressed rank $\wrtlaeryl$, and $\beta$, $\alpha=1-\beta$ are the linear interpolation factors which are given by
\begin{align*}
    \beta &= \frac{{ \localloss\brackets{\wrtlaeryl} - \localloss\brackets{\wrtlaeryl^{(l)}} }}{{\localloss\brackets{\wrtlaeryl^{(h)}} - \localloss\brackets{\wrtlaeryl^{(l)}}}},\\
    \alpha&=1-\beta.
\end{align*}
The motivation for this approximation is derived from observing the first-order Taylor series expansion of the change in the output due to weight perturbation:
\begin{equation*}
\begin{alignedat}{1}
    &\norm{f\brackets{\wlayerlv}-f\brackets{\wlayerlv+\Delta\wlayerlv}}_F^2    \approx &\norm{\Delta{\wlayerlv}^T\matsym{J}}_F^2+\mathcal{O},
\end{alignedat}
\end{equation*}
where $\matsym{J}\triangleq \frac{\partial\net\brackets{\wlayerlv}}{\partial\wlayerlv}$ is the neural network Jacobin matrix w.r.t. the weight vector $\wlayerlv$. We claim that under a small perturbation of the weight vector $\wlayerlv$ $\brackets{\text{e.g., }\norm{\Delta\wlayerlv}_F\leq \epsilon}$, and since the Jacobian matrix $\matsym{J}$ is independent of $\Delta\vectorsym{\wlayerlv}$ we have that
\begin{equation*}
    \norm{f\brackets{\vectorsym{\wlayerlv}}-f\brackets{\vectorsym{\wlayerlv}+\Delta\vectorsym{\wlayerlv}}}_F^2    \propto \norm{\Delta\vectorsym{\wlayerlv}}_F^2.
\end{equation*}
The interpolation method in~\eqref{eq:linear_interpolation} allows us to significantly reduce the number of required inference passes of the model while obtaining a close approximation to the actual error induced by each rank.
This is supported by empirical evidence as shown in Figure~\ref{fig:interpolation}, which shows the proximity of approximation given by the interpolated calculation compared to the actual error for each rank for an example layer in the DeiT-T model.

\section{Additional Low-Rank Ablation Results}
\label{apx:additional-results-lr}

Figure~\ref{fig:ablation-lr-vitb} and Figure~\ref{fig:ablation-lr-deitt} present a comparison between {\name} and the same algorithmic framework, considering only mixed-precision options for various weight compression rates applied to ViT-B and DeiT-T. 
Activations are uniformly quantized to 4-bit.
For larger models like ViT-B, the results emphasize the importance of integrating low-rank options into the optimization search, particularly at high compression rates, where they lead to significant accuracy improvements.
In contrast, for the smaller DeiT-T model, introducing low-rank compression into the mixed-precision search has little effect. This observation supports the idea that low-rank approximation tends to be more effective than quantization when applied to large matrices.

\section{{\name} Compression Solutions}
\label{apx:compression-solutions}
\begin{figure*}[t]
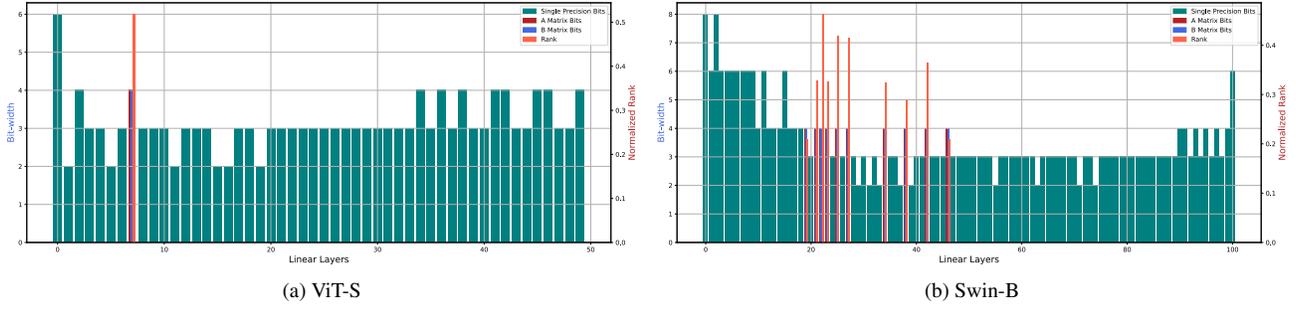

    \centering
    \begin{subfigure}{0.49\linewidth}
        \centering
        \includesvg[width=1.0\linewidth]{images/solution_illustration/solution_vit-s_3W4A.svg}
    \caption{ViT-S}
    \label{fig:comp-solution_vit_s}
    \end{subfigure}
    \begin{subfigure}{0.49\linewidth}
        \centering
        \includesvg[width=1.0\linewidth]{images/solution_illustration/solution_swin-b_3W4A.svg}
    \caption{Swin-B}
    \label{fig:comp-solution_swin_b}
    \end{subfigure}
    \caption{Compression solution for ViT-S and Swin-B with 3-bit average weights and 4-bit activations. Each layer is assigned with either a single precision bit-width or decomposed into low-rank matrices $A$ and $B$.}
\end{figure*}

Figure~\ref{fig:comp-solution_vit_s} and Figure~\ref{fig:comp-solution_swin_b} illustrate the compression solutions for ViT-S and Swin-B, using an average of 3-bit weights and 4-bit activations. Each layer is assigned a single precision bit width (selected from the set of options $\mathcal{B}=[2, 3, 4, 6, 8]$) or is decomposed into low-rank matrices $A$ and $B$. 
Notably, Swin-B, a substantially larger model, employs a low-rank approximation in 10 layers, while ViT-S, being much smaller, applies it to only one layer. This highlights the effectiveness of low-rank approximation over quantization, particularly for large matrices.
\begin{figure}[t]
    \centering    
    \includesvg[width=\linewidth]{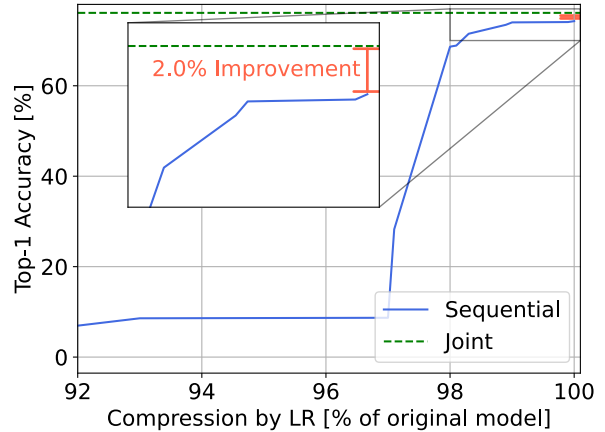}
    \caption{Comparison between our joint optimization approach (green-dashed line) and sequentially applying a percentage of compression using low-rank and the rest of the compression using quantization (blue line), to achieve weights memory equivalent to 3-bit, on DeiT-B.}
    \label{fig:sequential}
\end{figure}

\section{Candidates Solution Pareto Frontier Illustration}\label{apx:pareto-ill}

\begin{figure*}[t]
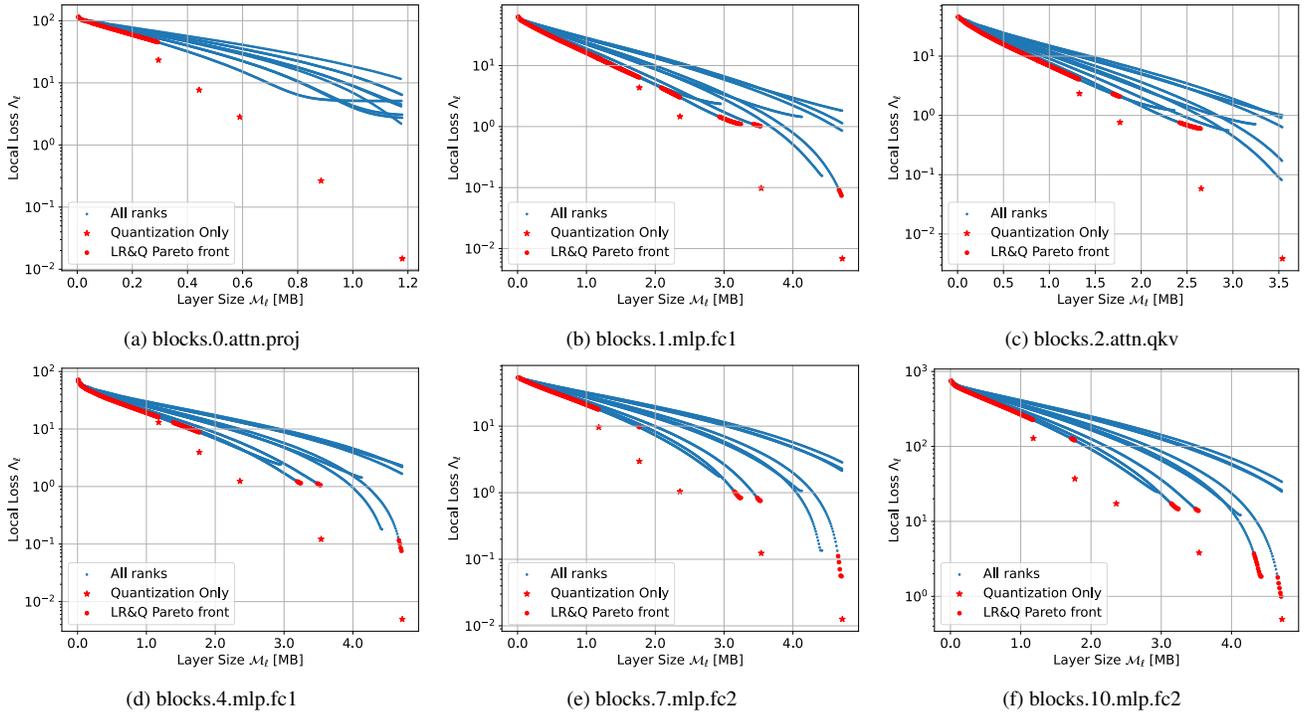

    \centering
    \begin{subfigure}{0.33\textwidth}
        \centering
        \includesvg[width=\linewidth]{images/pareto_illustrations/vit_s_16_blocks.0.attn.proj_pareto.svg}
        \caption{blocks.0.attn.proj}
    \end{subfigure}
    \hfill
    \begin{subfigure}{0.33\textwidth}
        \centering
        \includesvg[width=\linewidth]{images/pareto_illustrations/vit_s_16_blocks.1.mlp.fc1_pareto.svg}
        \caption{blocks.1.mlp.fc1}
    \end{subfigure}
    \hfill
    \begin{subfigure}{0.33\textwidth}
        \centering
        \includesvg[width=\linewidth]{images/pareto_illustrations/vit_s_16_blocks.2.attn.qkv_pareto.svg}
        \caption{blocks.2.attn.qkv}
    \end{subfigure}
    
    \begin{subfigure}{0.33\textwidth}
        \centering
        \includesvg[width=\linewidth]{images/pareto_illustrations/vit_s_16_blocks.4.mlp.fc1_pareto.svg}
        \caption{blocks.4.mlp.fc1}
    \end{subfigure}
    \hfill
    \begin{subfigure}{0.33\textwidth}
        \centering
        \includesvg[width=\linewidth]{images/pareto_illustrations/vit_s_16_blocks.7.mlp.fc2_pareto.svg}
        \caption{blocks.7.mlp.fc2}
    \end{subfigure}
    \hfill
    \begin{subfigure}{0.33\textwidth}
        \centering
        \includesvg[width=\linewidth]{images/pareto_illustrations/vit_s_16_blocks.10.mlp.fc2_pareto.svg}
        \caption{blocks.10.mlp.fc2}
    \end{subfigure}

    \caption{Inra-layer Pareto optimization results illustration for the ViT-S model.}
    \label{fig:pareto-ill}
\end{figure*}

Figure~\ref{fig:pareto-ill} illustrates the Pareto search solution for several layers of the ViT-S model.
The figure for each layer presents the local error and memory footprint values of each compression option from the full options space, meaning, for each of the bit-width and the rank options.
The red points mark the Pareto frontier, i.e., options that are not worse than any other option in both objectives (see definition in Section~\ref{sec:intra-layer}).
The solution yields the set of potentially optimal compression candidates for the layer, which are then used in the global mixed rank and precision search (Section~\ref{sec:inter-layer}) to obtain a unified compression solution for the model.
It can be observed from the illustration how the options space is massively reduced, making the search problem much more feasible.



\section{Sequential vs. Joint Low-Rank and Quantization}
\label{apx:seq-vs-joint}
To motivate our approach of a combined low-rank and quantization problem and joint optimization, we experimented, attempting to compress a DeiT-B model's weights to a target size in a sequential manner---first, by applying percentages of the compression using low-rank only and compressing the rest using quantization.
We compared this approach to our joint optimization approach in Figure~\ref{fig:sequential}.
As can be seen, applying the two techniques sequentially does not perform well when required to compress more than 3\% of the weights using low-rank, compared to our joint approach, which exploits the benefits from both techniques. 
This allows to achieve improved performance, while also removing the necessity of exploring the optimal low-rank percentage of the compression.

\section{Algorithmic Workflow}
\label{apx:workflow}

The complete algorithmic flow of \name{} is described in Algorithm~\ref{alg:workflow}.

\begin{algorithm*}[t]
   \caption{Mixed Low-Rank and Quantization (\name{}) Optimization Workflow}
   \label{alg:workflow}
    \begin{algorithmic}[1]
       \STATE {\bfseries Input:}
        Representative dataset $\mathcal{D}$, Pre-trained model $\net$ with weights $\wlayerl$ for the $\ell^{th}$ linear layer, Weight Memory constraint $\psi_{WMS}$. $\quad\quad\quad\quad\quad\quad\quad\quad\quad\quad\quad\quad\quad\quad\quad\quad\quad\quad\quad\quad\quad\quad\quad\quad\quad\quad\quad\quad\quad\quad\quad\quad\quad\quad\quad\quad\quad\quad\quad\quad\quad\quad\quad\quad\quad\quad\quad\quad\quad\quad\quad\quad\quad\quad\quad\quad$
       \phase{Intra-Layer Search}
       \FOR{$\ell=1$ to $L$}
           
           \STATE Compute the per-channel diagonal Hessian matrix bound  $\matsym{Q}_{\ell}$~\eqref{eq:quant_aware_decomp}.
           
           \STATE Perform Hessian-aware SVD decomposition: $\matsym{U}_{\ell},\matsym{\Sigma}_{\ell},\matsym{V}_{\ell}=svd\brackets{\matsym{Q}_{\ell}\matsym{W}_{\ell}}$.
           
           \STATE Compute low-rank matrices: $\matsym{A}_{\ell}= \matsym{Q}_{\ell}^{-1}\matsym{U}_{\ell}$, $\matsym{B}_{\ell} = \matsym{\Sigma}_{\ell}\matsym{V}_{\ell}^T$.
           
           \STATE Determine quantization parameters $\qpa$ and $\qpb$ by minimizing \eqref{eq:threshold_upper_bound_a} and \eqref{eq:threshold_upper_bound_b}, respectively.
           
           \STATE Determine quantization threshold $\qpw$ for $\wlayerl$ using Hessian Mean Square Error in \eqref{eq:hmse-loss}.
           
           \STATE {\bfseries Pareto Set Search:}
           \FORALL{compression options $\wrtlaeryl\in\mathcal{W}_{\ell}$~\eqref{eq:compressed_operation}}
               
               
               \STATE Compute loss $\localloss$ using \eqref{eq:problem_intra_final}
               
               \STATE Compute memory footprint $\mathcal{M}_{\ell}$ using \eqref{eq:memory_constraint}
           \ENDFOR
           \STATE Find Pareto optimal solutions: $\mathcal{P}_{\ell}$ 
       \ENDFOR \\
       \phase{Inter-Layer Search}
       \STATE Compute the global SQNR error~\eqref{eq:global_optimization} per candidate in $\mathcal{P}_{\ell}$ for all layers.
       \STATE Formulate Integer-Linear Programming (ILP) problem~\eqref{eq:ilp}.
       \STATE Solve ILP to find optimal compression configuration $\mathcal{S}$.
       \STATE Compute activation mixed precision allocation using \eqref{eq:act_mp_close}.\\
       \phase{Low-Rank Aware Adaptive Rounding Search}
       \STATE Apply quantization corrections (from another method e.g. RepQ~\cite{li2023repq}, ERQ \cite{10839431erq})
       \STATE Apply Low-Rank-aware adaptive rounding on quantized weights. \\
    \end{algorithmic}
\end{algorithm*}

\end{document}